\documentclass[journal]{IEEEtran}
\ifCLASSINFOpdf
\else
\fi
\usepackage{booktabs} 
\usepackage{multirow} 
\usepackage{makecell} 
\usepackage{float} 
\usepackage{amsmath} 
\usepackage{amsfonts} 
\usepackage{graphicx} 
\usepackage[caption=false, font=footnotesize]{subfig} 
\usepackage{url}
\usepackage[ruled,linesnumbered]{algorithm2e} 
\SetKwInput{KwInput}{Input}
\SetKwInput{KwOutput}{Output}
\SetKwComment{Comment}{/* }{ */}
\usepackage[dvipsnames]{xcolor} 
\usepackage{stfloats} 

\begin{document}
%
\title{SPEC: Summary Preference Decomposition for Low-Resource Abstractive Summarization}
%
%
%

\author{Yi-Syuan~Chen,
        Yun-Zhu~Song,
        and~Hong-Han~Shuai,~\IEEEmembership{Member,~IEEE}
\IEEEcompsocitemizethanks{
\IEEEcompsocthanksitem This work is supported in part by the National Science and Technology Council (NSTC) of Taiwan under the grants NSTC-109-2221-E-009-114-MY3, NSTC-111-2221-E-A49-164 and NSTC-1111-2221-E-001-021. \\ 
\indent Y.-S. Chen, Y.-Z. Song and H.-H. Shuai are with the Department of Electronics and Electrical Engineering, National Yang Ming Chiao Tung University, No. 1001, University Road, Hsinchu 300, Taiwan. E-mail: yschen.ee09@nycu.edu.tw, yzsong.ee07@nycu.edu.tw, hhshuai@nycu.edu.tw. \\
\indent This paper has supplementary downloadable material available at http://ieeexplore.ieee.org., provided by the author. The material includes additional supporting experiments for the main paper. Contact yschen.ee09@nycu.edu.tw for further questions about this work.
}
}

%
%

\markboth{Manuscript for IEEE Transactions on Audio, Speech and Language Processing (TASLP)}%
{Shell \MakeLowercase{\textit{et al.}}: Bare Demo of IEEEtran.cls for IEEE Journals}
%



\maketitle
\newcommand\TASLPabbrev{SPEC}
\newcommand\AAAIabbrev{MTL-ABS}

\begin{abstract}
Neural abstractive summarization has been widely studied and achieved great success with large-scale corpora. However, the considerable cost of annotating data motivates the need for learning strategies under low-resource settings. In this paper, we investigate the problems of learning summarizers with only few examples and propose corresponding methods for improvements. First, typical transfer learning methods are prone to be affected by data properties and learning objectives in the pretext tasks. Therefore, based on pretrained language models, we further present a meta learning framework to transfer few-shot learning processes from source corpora to the target corpus. Second, previous methods learn from training examples without decomposing the \textit{content} and \textit{preference}. The generated summaries could therefore be constrained by the preference bias in the training set, especially under low-resource settings. As such, we propose decomposing the contents and preferences during learning through the parameter modulation, which enables control over preferences during inference. Third, given a target application, specifying required preferences could be non-trivial because the preferences may be difficult to derive through observations. Therefore, we propose a novel decoding method to automatically estimate suitable preferences and generate corresponding summary candidates from the few training examples. Extensive experiments demonstrate that our methods achieve state-of-the-art performance on six diverse corpora with 30.11\%/33.95\%/27.51\% and 26.74\%/31.14\%/24.48\% average improvements on ROUGE-1/2/L under 10- and 100-example settings.
\end{abstract}


%
\IEEEpeerreviewmaketitle
\section{Introduction}
\label{sec:introduction}
%
%
%
%


\IEEEPARstart{N}{eural} abstractive summarization has drawn much attention due to its wide applications, \textit{e.g}., news headline generation~\cite{Song_Shuai_Yeh_Wu_Ku_Peng_2020}, consumer health question summarization~\cite{ben-abacha-demner-fushman-2019-summarization}, conference talk summarization~\cite{lev-etal-2019-talksumm}, or livestream transcript summarization~\cite{cho2021streamhover}. With large annotated corpora, the results of neural abstractive summarization could generally approach the ground truths~\cite{liu-etal-2022-brio,song-etal-2022-improving}. However, preparing the annotations, which requires experts with highly sophisticated skills such as information selection, reorganization, and distillation, is labor-intensive and expensive, especially when the target corpus is of an unusual type or related to professional fields such as product reviews~\cite{keung-etal-2020-multilingual}, scientific papers~\cite{cachola-etal-2020-tldr}, or legislative documents~\cite{sharma-etal-2019-bigpatent}.

Therefore, to alleviate annotation efforts, the \textit{Low-Resource Abstractive Summarization} has emerged as an important technology that aims to improve generation quality with limited annotated target examples. In the literature, transfer learning has been utilized to tackle low-resource problems in many applications, including abstractive summarization~\cite{radford2019language,DBLP:journals/corr/abs-1905-08836,lewis-etal-2020-bart,pmlr-v119-zhang20ae,fabbri-etal-2021-improving}. In transfer learning, models are learned with some \textit{pretext tasks} before adapting to the downstream tasks~\cite{10.1007/978-3-319-46487-9_40,gidaris2018unsupervised}. There are two common paradigms for creating such pretext tasks, \textit{i.e.}, \textit{self-supervised learning} and \textit{intermediate-finetuning}. Specifically, in self-supervised methods,  pseudo annotations are crafted by some heuristics on unannotated data to train models in a supervised learning manner~\cite{radford2019language,DBLP:journals/corr/abs-1905-08836,lewis-etal-2020-bart,pmlr-v119-zhang20ae}. On the other hand, for intermediate finetuning methods, task-related annotated datasets are used to finetune language models before finetuning on the target task~\cite{fabbri-etal-2021-improving}. Although transfer learning methods are usually effective, they are prone to be affected by the data properties or learning objectives in the pretext tasks~\cite{Chiang_Lee_2022,Wang_2019_CVPR}. Different from transfer learning, meta learning provides a framework that transfers \textit{task-level} knowledge instead of \textit{data-level} one. It simulates the train-test process with few examples during training and optimizes the models to reduce the simulated test errors~\cite{pmlr-v70-finn17a}. In such schemes, the behaviors of learning with few examples are specifically considered. The ability of "learning to learn" allows the model to adapt to unseen tasks quickly. 

Nevertheless, two challenges arise in designing such a meta learning framework for abstractive summarization. 1) \textit{Expensive computation.} Meta learning involves the computation of second derivatives, which is expensive for summarization models due to the typically large model size. 2) \textit{Meta-dataset construction.} Meta learning requires crafted tasks for learning. In applications such as image classification~\cite{Sun_2019_CVPR}, tasks can be directly defined as different class combinations in a single dataset, which is not suitable for abstractive summarization. As an alternative, we could leverage different corpora as different tasks. However, this further introduces multi-tasking and domain generalization problems~\cite{pmlr-v139-triantafillou21a}.

In this paper, to solve these two problems, we propose the \textbf{M}eta-\textbf{T}ransfer \textbf{L}earning for low-resource \textbf{ABS}tractive summarization (\AAAIabbrev). Specifically, to address the first challenge, we propose leveraging adapters with a limited number of parameters to perform meta learning with pretrained language models. With respect to the second challenge, we formulate the meta-dataset construction as a corpus-choice problem and further investigate the meta-transferability to obtain empirical rules for constructing meta-datasets.

\begin{table*}[t]
\scriptsize
\renewcommand\arraystretch{1.0}
\caption{Generation cases from models learned with different examples. Phrases extracted from the article are marked in italics.}\smallskip
\label{tb:pilot}
\centering
\begin{tabular}{p{0.47\textwidth}p{0.47\textwidth}}
\toprule
\textbf{Article:} The man died in Inverness on 27 October this year. The Police Investigations and Review Commissioner (Pirc), Kate Frame, has been asked to scrutinise the initial police response to the man's call. Police Scotland said it was "fully engaging" with the investigation and awaited its findings. A spokesman for Pirc said: "The Crown Office and Procurator Fiscal Service (COPFS) has instructed the Police Investigations and Review Commissioner to undertake an investigation into the initial police response to a call from a 72-year-old man who was later found dead at a sheltered housing complex in Inverness. "A report on the commissioner's findings will be submitted to the COPFS in due course." &
\textbf{Article:} Coventry University's Scarborough campus has been built on the town's former Weaponness Park and Ride site. About 200 students have begun courses at the site, though it is expected to eventually be home to more than 2,000 students. The building, which includes engineering and science labs, a mock law court and a library, is part of a £50m sports and education facility. Professor Craig Gaskell said: "Launching our new state-of-the-art building is a huge milestone for us and demonstrates our commitment to Scarborough and the Yorkshire coast area." A new University Technical College has been built nearby and Scarborough Athletic FC's new 2,000-seater stadium is also under construction on the site. Coventry University also has a campus near London's Liverpool Street Station and recently announced it will open a campus in Dagenham in September 2017. \\
\midrule
\textbf{Ground Truth:} An investigation has been launched into the police response \textit{to a call from a 72-year-old man who was later found dead in a sheltered housing complex}. &
\textbf{Ground Truth:} A new £14m university campus has opened in Scarborough. \\
\textbf{Article-Summary Similarity:} 0.463 (highest-level) &
\textbf{Article-Summary Similarity:} 0.117 (middle-level) \\
\midrule
\textbf{Learned with 100 examples in highest similarity level:} \textit{A 72-year-old man} who was \textit{found dead at a sheltered housing complex in Inverness} has been asked to \textit{scrutinise the initial police response to a call}. &
\textbf{Learned with 100 examples in highest similarity level:} The University of Coventry has opened a new campus in Scarborough that will house \textit{more than 2,000 students}. \\
\textbf{ROUGE-1/2/L:} 78.57/55.56/46.43 &
\textbf{ROUGE-1/2/L:} 57.14/23.08/35.71 \\
\midrule
\textbf{Learned with 100 examples in middle similarity level:} Police Scotland have been asked to scrutinise the response to the death of a man who was \textit{found dead at a sheltered housing complex}. &
\textbf{Learned with 100 examples in middle similarity level:} A new campus at the University of Scarborough has been opened in Scarborough. \\
\textbf{ROUGE-1/2/L:} 57.69/28.00/53.85 &
\textbf{ROUGE-1/2/L:} 72.73/30.00/63.64 \\
\bottomrule
\end{tabular}
\end{table*}

\AAAIabbrev~benefits from the meta learning framework that utilizes diverse existing corpora. However, in our further investigations, we found that the generated summaries could inherit the summary properties of source corpora. This is an inevitable phenomenon commonly existing in transfer or meta learning methods. To illustrate this phenomenon, we demonstrate the generation results on testing examples with different article-summary similarity levels. Specifically, we first pick two sets of 100 examples from the XSum dataset~\cite{narayan-etal-2018-dont} based on the lexical similarity~\cite{lin-2004-rouge} between the articles and ground truth summaries. One set has the highest similarity level in the dataset, and the other one is in the middle level (centered at the top-50\%). The similarity between articles and summaries suggests different summary properties. High similarity indicates the summaries are more extractive, while low similarity indicates the summaries are more abstractive. Next, we initialized two models with the BART~\cite{lewis-etal-2020-bart} and finetuned them respectively with the two sets. Table~\ref{tb:pilot} shows the generation results from the two models on two unseen testing examples with different similarity levels, together with the ROUGE scores between the generated and ground truth summaries. The results manifest that the performance can be highly affected by the training examples under the low-resource setting, as the ROUGE scores vary drastically between the two models. This is because the two models may learn the specific bias from the training set, and the performance is better if the downstream requirements meet the bias. However, the generation results for both sets are fluent and concise for expressing the content of the article. The one trained by the high similarity set tends to use phrases extracted from the article such as \textit{"a 72-year-old man"} and \textit{"scrutinise the initial police response to a call"} for the high similarity testing example, while using the phrase \textit{"more than 2,000 students"} in the middle similarity testing example. Nevertheless, the semantics of summaries from the two models are generally similar. Therefore, we argue that the data with different article-summary similarities may offer similar supervisions on the contents for model learning, while typical training methods could constrain the generations and thus affect the performance.

Therefore, we further improve the \AAAIabbrev~by two key facts. First, previous methods learn from training examples without decomposing the \textit{content} and \textit{preference}. The content refers to the main idea conveyed by the summary, while the preference refers to the manifestation of the content. Second, previous methods aim to extract useful knowledge from existing corpora or massive general texts~\cite{fabbri-etal-2021-improving}. However, the unannotated documents from the target domain are not explored, which may provide knowledge that is distinct to the domain. Nevertheless, these two ideas raise three new challenges. 1) \textit{Compatibility with pretrained models.} It is typical to use pretrained language models as the initialization before transfer or meta learning. To introduce the preference signals, new modules should be added. However, an inappropriate architecture design would deteriorate pretrained knowledge~\cite{pmlr-v97-houlsby19a}. 2) \textit{Distinct knowledge sources.} How to properly transfer knowledge from unannotated documents and meanwhile incorporate other knowledge sources with summaries remains an open problem. 3) \textit{Preference Assignment.} It is non-trivial for users to specify required preferences during inference because preferences are essentially scalars that may not be easy to be obtained solely from observations.
 
Therefore, to solve these three problems, we propose the \textbf{S}ummary \textbf{P}reference d\textbf{EC}omposition (\TASLPabbrev). For the first problem, based on \AAAIabbrev, we propose decomposing the content and preference through the parameter modulation~\cite{Perez_Strub_de_Vries_Dumoulin_Courville_2018} on the adapters. For the second problem, we extend the solution for the first problem to use two distinct sets of adapters for transfer and meta learning methods separately. The sets of adapters are learned alternatively at each step, which is reminiscent of multi-task learning. For the third problem, we propose a clustering-based decoding method to estimate suitable preferences from few examples and then generate summary candidates for users to select. The contributions are summarized as follows.

\begin{itemize}
\item We propose the \AAAIabbrev, which utilizes adapters and crafted meta-datasets to leverage task-level knowledge from multiple sources for improvements under low-resource settings.

\item Based on the \AAAIabbrev, we further propose the \TASLPabbrev, which decomposes the content and preference during learning and enables control over preferences during inference. Moreover, a novel decoding method is proposed to generate suitable summary candidates for the target application automatically.

\item Extensive experiments demonstrate that the proposed methods achieve state-of-the-art performance on 6 diverse corpora with 30.11\%/33.95\%/27.51\% and 26.74\%/31.14\%/24.48\% average improvements on ROUGE-1/2/L under 10- and 100-example settings. In addition to the superior performance, the philosophy of decomposing preferences and contents could provide an inspirational viewpoint beyond conventional learning schemes.
\end{itemize}

\section{Related Works}
\label{sec:related_work}
\subsection{Low-Resource Learning Strategies}
To mitigate the problems of data scarcity, transfer learning has been widely adopted in different applications~\cite{9134370}. A recent line of studies proposes different self-supervised objectives to learn language models that encapsulate ground knowledge shared across NLP tasks~\cite{devlin-etal-2019-bert,NEURIPS2019_dc6a7e65,JMLR:v21:20-074,lewis-etal-2020-bart, pmlr-v119-zhang20ae}, which enables faster convergence and better accuracy on downstream tasks, including abstractive summarization. For example, T5~\cite{JMLR:v21:20-074} leverages BERT-style~\cite{devlin-etal-2019-bert} objectives on encoder-decoder architectures to facilitate downstream generation tasks. BART~\cite{lewis-etal-2020-bart} proposes several denoising objectives also for encoder-decoder architectures, including token masking, token deletion, text infilling, sentence permutation, and document rotation, which additionally explores sentence-level knowledge. PEGASUS~\cite{pmlr-v119-zhang20ae} further proposes Gap Sentence Generation (GSG) objective to improve the conditional generation ability of the decoder, which is highly related to abstractive summarization. It selects and masks important sentences in an article according to the ROUGE scores between the sentence and the rest of the article, and the decoder learns to reconstruct the masked sentences. In addition to self-supervised learning, models could also benefit from other supervised tasks related to the target one. Specific for abstractive summarization, PreSumm~\cite{liu-lapata-2019-text} proposes a Transformer-based framework with the encoder first initialized with the BERT~\cite{devlin-etal-2019-bert} and then finetuned on an extractive summarization objective. \cite{magooda2021exploring} incorporates different tasks, including extractive summarization, language modeling, concept detection, and paraphrase detection, to explore multi-task learning for low-resource improvements. WikiTransfer~\cite{fabbri-etal-2021-improving} leverages Wikipedia documents, assigning the first few sentences as summaries and the rest as articles. The collected data are then binned into five levels according to the ROUGE scores between articles and summaries. Under the assumption that the ROUGE performance of extractive oracles is known for the target corpus, it filters out Wikipedia data that are located in the same bin with the target corpus and uses them to finetune the base model. Compared with the proposed SPEC, the training of the above methods is typically conducted on general domain data. When encountering data in a specific domain, the learned knowledge may become an obstacle to adapting to the new tasks.

To transfer pretrained models to downstream tasks, it is common to add some task-specific layers on top and finetune the model in part or in full~\cite{liu-etal-2019-multi, magooda2021exploring}. However, this strategy is often inefficient for abstractive summarization. The output layers of summarizers typically contain a large number of parameters since the output dimension is set according to the vocabulary size. Moreover, the output layers are often tied with input layers for learning efficiency~\cite{inan2016tying}. As such, full retraining with adequate annotated examples is still required when encountering new tasks. To alleviate such circumstances, \cite{pmlr-v97-houlsby19a} proposes a compact \textit{adapter} module to transfer knowledge from BERT~\cite{devlin-etal-2019-bert} to natural language understanding tasks. Each BERT layer is inserted with few adapter modules, and only the adapter modules are set to be learnable. Similarly, \cite{pmlr-v97-stickland19a} transfers the knowledge in BERT with Projected Attention Layers (PALs) for multi-task natural language understanding, which is a multi-head attention layer residually connected to the base model.

Another line of research for the low-resource solution is meta learning. Different from typical transfer learning, meta learning aims to transfer knowledge that is superordinate to the data-level. For example, MAML~\cite{pmlr-v70-finn17a} is proposed to learn from few-shot tasks. In each task, the model first learns on a small \textit{support set} and then tests on a small \textit{query set}. The test errors of tasks are collected for optimization. Such a learning objective allows models to learn to generalize to the query set given the support set, which is specifically designed for low-resource settings. Meta learning has been shown to perform well in a variety of NLP tasks, including machine translation~\cite{Li_Wang_Yu_2020,gu-etal-2018-meta}, dialog systems~\cite{qian-yu-2019-domain,madotto-etal-2019-personalizing}, relational classification~\cite{obamuyide-vlachos-2019-model}, semantic parsing~\cite{guo-etal-2019-coupling}, emotion learning~\cite{zhao-ma-2019-text}, and natural language understanding~\cite{dou-etal-2019-investigating}. Based on the aforementioned research lines, the proposed \TASLPabbrev~aims to coordinate the transfer learning, meta learning, and efficient transfer modules to improve low-resource performance.

\subsection{Learning with Preferences}
Generating summaries with preferences is a long-standing problem. Specifically, the preferences could be embodied in various perspectives. For examples, \cite{fan-etal-2018-controllable, chan-etal-2021-controllable} consider lexical properties such as length or covered entities. \cite{jin-etal-2020-hooks} considers writing styles that could invoke feelings such as humor or romance. \cite{tan-etal-2020-summarizing} summarizes documents with user provided aspects. \cite{gao2020preference} learns models based on summary comparisons from users to serve the preferences. \cite{DBLP:journals/corr/abs-1909-08593,NEURIPS2020_1f89885d,nguyen-etal-2022-make} instead learn to generate summaries that are better according to the human consensus, which is a more general preference. These methods could be categorized into two classes according to whether we could inject user preferences during inference, \textit{i.e.}, \textit{inference-time control} or \textit{non-inference-time control}~\cite{cao-wang-2021-inference}. Inference-time control aims to generate flexible summaries given the conditions that are not defined to serve specific users, \textit{e.g.}, lexical and semantic properties. In other words, the models are built in the \textit{one-for-all} manner and can generate summaries according to the downstream requirements. For non-inference-time control, the preferences are usually closely related to individuals, and the problems are typically formulated under the concept of \textit{human-in-the-loop}~\cite{nguyen-etal-2022-make}, which allows humans to participate model learning process through actively annotating examples or providing feedback. Specifically, these methods are often designed based on reinforcement learning~\cite{gao2020preference,DBLP:journals/corr/abs-1909-08593,NEURIPS2020_1f89885d,nguyen-etal-2022-make}. The summarizers are optimized to increase rewards specifying the matching degree of preferences for summaries. The rewards could be provided by online human annotations~\cite{sokolov-etal-2016-learning}, reward models learned from human feedbacks~\cite{DBLP:journals/corr/abs-1909-08593, NEURIPS2020_1f89885d}, or in hybrid~\cite{gao2020preference, nguyen-etal-2022-make}. The models in these methods are typically \textit{one-for-one} to meet distinct requirements.

Our methods are close to the inference-time control methods but are different from previous works in three ways. First, we focus on the low-resource settings, where only few training examples are available. In such scenarios, data bias during learning could greatly affect the downstream generation results. Therefore, we propose to disentangle the contents and preferences for better leveraging source data and satisfying the target requirements. Second, instead of considering exact styles such as romantic or humorous, we focus on the statistical relations between the articles and summaries such as ROUGE~\cite{lin-2004-rouge}, extractive degree~\cite{grusky-etal-2018-newsroom}, novel word ratio, and compression ratio. These preferences jointly exist in corpora in different proportions. Therefore, our preference-aware summarizer aims to learn these preferences disentangled with the contents for better generalization on target corpora. Third, previous works leave the flexibility for generating summaries with some preferences but do not specify how users could assign the preferences. However, the preferences are essentially some scalars that could be difficult to obtain from observations. In contrast, we further propose a novel decoding method to learn appropriate preferences directly from data without the requirement of specifying by heuristics. Although our learning is conducted in a one-for-all manner, the proposed decoding process creates the one-for-one prediction results. Therefore, our methods are still evaluated with human-written summaries in the target domain, while previous works leverage model-based scorers or humans to quantify the improvement toward specified preferences.

\section{Meta-Transfer Learning for Abstractive Summarization}
The bi-level optimization in meta learning involves computing the Hessian matrix for all learnable parameters, and the computation and memory footprint would increase quadratically with the number of parameters. This problem is especially severe for summarization models that are typically complex. Therefore, we introduce parameter-efficient adapters to take advantage of meta learning with limited parameter counts. Moreover, since there is no clear strategy to create meta-datasets for abstractive summarization, we study the meta-transferability and propose several empirical rules for constructing meta-datasets. The following subsections sequentially present the architectures, meta-dataset construction, and a basic learning method.

\subsection{Base Model}
Based on the use of the Transformer~\cite{NIPS2017_3f5ee243}, our base architecture is mainly comprised of self-attention (SA) and cross-attention (CA) layers, while self-attention and cross-attention layers are composed of multi-headed attention (MHA), feed-forward (FF), and layer normalization (LN) layers. The multi-headed attention layer operates as follows:
\begin{equation}
    \text{A}(h,\tilde{h}) = softmax(\frac{(W_q h) \cdot (W_k \tilde{h})^T}{\epsilon})(W_v \tilde{h}),
\end{equation}
\begin{equation}
    \text{MHA}(h,\tilde{h}) = concat(\text{A}_1(h, \tilde{h}), \text{A}_2(h, \tilde{h}), ...)W_o,
\end{equation}
where $h, \tilde{h}$ are intermediate hidden states, $W_q, W_k, W_v, W_o$ are learnable parameters, and $\epsilon$ is a normalization scalar. The self-attention and cross-attention layers can then be written as:
\begin{equation}
\begin{split}
    \text{SA}(h) & = \text{LN}(\text{FF}(\text{MHA}(h, h)) + h), \\
    \text{CA}(h, h_{enc}) & = \text{LN}(\text{FF}(\text{MHA}(h,h_{enc})) + h),
\end{split}
\end{equation}
where $h_{enc}$ represents the final encoder hidden states. Based on the self-attention and cross-attention layers, the encoder (ENC) and decoder (DEC) layers can be expressed as:
\begin{equation}
\begin{split}
    \text{ENC}(h) = \text{LN}(\text{FF}( & \text{SA}(h)) + \text{SA}(h)), \\
    \text{DEC}(h, h_{enc}) = \text{LN}(\text{FF}( & \text{CA}(\text{SA}(h), h_{enc})) + \\
    & \text{CA}(\text{SA}(h), h_{enc})).
\end{split}
\end{equation}
We use BART~\cite{lewis-etal-2020-bart}, a sequence-to-sequence pretrained Transformer, to initialize both the encoder and decoder.

\subsection{Adapters}
\label{subsec:adapters}
Given the pretrained Transformer base model, we aim to limit the learnable parameters during meta learning to alleviate computation issues. A naive approach is to meta-learn only part of the layers to control the model complexity. However, such a scheme would greatly affect the pretrained structures and further degrade the performance. Therefore, we propose leveraging additional modules for this purpose. To introduce modules that are compatible with the pretrained model without degrading the performance, we adopt parameter-efficient adapters~\cite{pmlr-v97-houlsby19a}. The adapter is a bottlenecked feed-forward network consisting of a down-project layer (DP) and an up-project (UP) layer.
\begin{equation}
\begin{split}
    \text{DP}(h) & = W_{d} \cdot h, \\
    \text{UP}(\bar{h}) & = W_{u} \cdot \bar{h},
\end{split}
\end{equation}
where $h \in \mathbb{R}^m, \bar{h} \in \mathbb{R}^n $ are the intermediate hidden states, and $W_{d} \in \mathbb{R}^{n \times m}, W_{u} \in \mathbb{R}^{m \times n}$ are learnable parameters. The hidden size $n$ is set to be smaller than $m$. A skip-connection from input to output is established, which can prevent noisy initialization from interfering with training at an early stage. The adapter (ADA) can thus be expressed as:
\begin{equation}
    \text{ADA}(h) = \text{UP}(\text{ReLU}(\text{DP}(h))) + h.
\end{equation}
The adapters are used after each feed-forward layer to leverage the pretrained knowledge better.

\subsection{Meta-Datasets}
\label{subsec:meta_datasets}
In applications such as classification~\cite{Sun_2019_CVPR,obamuyide-vlachos-2019-model}, the tasks for meta learning can be directly defined as different combinations of classes. However, it is not applicable to abstractive summarization since there is no clear heterogeneity among data. Instead of randomly sampling data from a single corpus to construct tasks, we propose leveraging multiple corpora. Specifically, since the tasks are defined as data from different corpora, the problem thus becomes how to choose the source corpora. An inappropriate choice could cause a negative transfer problem and degrade performance. Therefore, the intuitive idea is to choose diverse source corpora. Meanwhile, each source corpus should possess as many identities to the target corpus as possible. This idea could be achieved with properly designed similarity criteria, where the performance of models should show a monotonous change with corpora chosen along the similarity rankings. Based on this idea, we consider the following hypotheses that may help in source corpora choice. 1) \textit{Semantics}: the document embedding similarity, 2) \textit{Word overlapping}: token cosine similarity, 3) \textit{Information coverage}: ROUGE recall, 4) \textit{Information density}: ROUGE precision, 5) \textit{Length}: absolute difference in token length between articles.

Given a target corpus, we rank the source corpora and create meta-datasets in different ranking sections for each criterion. Next, we conduct experiments to investigate the relation between performance and source corpora choice\footnote{The analysis and implementation are detailed in the Appendix.}. According to the results, we have the following observations. First, the statistical similarity is more important than semantic similarity. Second, corpora from the same domain may not always provide better knowledge. Third, rather than choosing source corpora that can cover all used words in the target corpus, source corpora should have a high information density. Fourth, source corpora with similar average article lengths are better than source corpora with different average article lengths. With the above observations, we use the average ranking of the following criteria: 1) cosine similarity, 2) ROUGE precision, and 3) article length to choose our source corpora for the meta-dataset.

\subsection{Bi-level Optimization}
\label{subsec:basic_learning_paradigm}
Given the adapter-enhanced summarizer, we propose a bi-level optimization to coordinate transfer and meta learning. Consider a pretraining corpus $C^{pre}$, a set of source corpora $\{ C^{src}_j \}$, and a target corpus $C^{tgt}$, we aim to leverage both $C^{pre}$ and $\{ C^{src}_j \}$ to improve the performance on $C^{tgt}$ containing only few annotated examples. Our summarizer comprises a base model and adapters, which are parameterized with $\theta$ and $\phi$, respectively. We first learn the base model on $C^{pre}$ for acquiring preliminary capability of summarizing. Given a specific document $x$ from $C^{pre}$, the base model produces a particular prediction sequence $y_{<t}=[y_1,...,y_{t-1}]$ at time $t$, and the probability to generate next token $y_t$ is written as $p(y_t|y_{<t},x,\theta)$. With the corresponding summary $[\widehat{y}_1,...,\widehat{y}_{N_y}]$, we optimize the base model to minimize the negative log-likelihood (NLL) as:
\begin{align}
    -\log p(y|x,\theta)  = -\sum^{N_y}_{t=1}\log p(\widehat{y}_t|y_{<t},x,\theta),
\end{align}
\begin{equation}
    -\log p(C^{pre}|\theta) = -\sum_{(x,y) \in C^{pre}}\log p(y|x,\theta).
\end{equation}

After the training of the base model, we insert adapters in the base model. To meta-learn the adapters, we sample examples without replacement from a set of source corpora $\{ C^{src}_j \}$ to create a collection of tasks $\{\mathcal{T}_\tau\}_{\tau=1}^M$ as meta-dataset $\mathcal{M}$. The source corpora are chosen according to the proposed criteria in Sec.~\ref{subsec:meta_datasets}. Each task $\mathcal{T}_\tau$ contains a task-training set $D^{tr}_\tau=\{ (x_i^\tau,y_i^\tau) \}^K_{i=1}$ and a task-testing set $D^{te}_\tau=\{ ({x'}_i^{\tau}, {y'}_i^{\tau})\}^K_{i=1}$. The number of tasks from different corpora is further balanced to avoid domain bias. At each meta-step, we sample tasks to form a meta-batch $\mathcal{B}=\{\mathcal{T}_\tau | \tau \in B\}$. In the inner loop optimization, we initialize the \textit{task-parameters} $\psi$ from the \textit{meta-parameters} $\phi$. We then optimize $\psi$ with the following objective on the task-training set $D^{tr}_\tau$: 
\begin{equation}
    -\log p(D^{tr}_\tau|\psi,\phi) = -\sum_{(x,y) \in D^{tr}_\tau}\log p(y|x,\psi,\phi).
\end{equation}
Through the optimization, $\psi$ is adapted to the specific task $\tau$ as $\psi_\tau$. Assuming that there is only one update step, the update can be expressed as:
\begin{equation}
    \psi_\tau \leftarrow \psi + \beta\nabla_\psi \log p(D^{tr}_\tau|\psi,\phi), 
\end{equation}
where $\beta$ is the inner loop learning rate. For the outer loop optimization, meta-parameters $\phi$ minimize the testing loss after adaptation with task-testing set $D^{te}_\tau$ in batch $\mathcal{B}$ as follows:
\begin{equation}
    -\log p(\mathcal{B}|\phi)=-\sum_{\tau \in B} \log p(D^{te}_\tau|D^{tr}_\tau,\psi_\tau,\phi).
\end{equation}
The meta-parameters $\phi$ is then updated by:
\begin{equation}
    \phi \leftarrow \phi + \alpha \nabla_\phi \log p(\mathcal{B}|\phi),
\end{equation}
where $\alpha$ is the outer loop learning rate. With such an update scheme, the meta-parameters $\psi$ learn to minimize the generalization errors after adaptation with a small number of examples. After the meta-transfer learning process, we use the few annotated examples on the target corpus to further finetune the summarizer for adaptation.

\section{Summary Preference Decomposition}
\label{sec:spec}
As discussed in the introduction, previous methods learn from training examples with the content and preference tied up. As such, the generated summaries could be constrained by the preference bias existing in the training set, which may not meet user requirements, especially under low-resource settings. Therefore, we propose decomposing the content and preference through parameter modulation~\cite{Perez_Strub_de_Vries_Dumoulin_Courville_2018}. Specifically, the parameters in the adapters are transformed according to the features of the given preferences, which makes the models conditioned on preferences. In other words, the models are different (with the adapter parameters) when the provided preferences are different. Since the preferences contain much information about the required targets, the models would learn to leverage these "shortcuts." As such, we could decompose the content and preference during learning and provide the flexibility to control the preferences during inference. The following subsections sequentially present the architectures, learning algorithms, and decoding methods.

\subsection{Definition of Preference}
\label{subsec:preference_definition}
In this work, we focus on the statistical relations between the articles and summaries for estimating the preferences. Considering an article-summary pair $(x, y)$, the preference, denoted by $p$, is constructed by the following metrics. a) \textit{ROUGE}~\cite{lin-2004-rouge}. We use ROUGE-1, ROUGE-2, and ROUGE-L to measure the lexical similarity between $x$ and $y$ at different granularities. b) \textit{Extractive diversity/coverage}~\cite{grusky-etal-2018-newsroom}. We use extractive diversity and coverage to quantify the extractive degree of the summary. c) \textit{Novel word ratio}. We compute the novel word ratio $|y-x|/|y|$ to explicitly measure the abstractive degree of the summary. d) \textit{Compression ratio}. We compute the token length ratio $|y|/|x|$ to constrain the model for generating summaries with appropriate information density. Other perceptive metrics, such as sentiment or popularity~\cite{Song_Shuai_Yeh_Wu_Ku_Peng_2020}, can also be potential preferences, but we use the aforementioned metrics for generality.

\subsection{Preference-Aware Summarizer}
\begin{figure}[t]
\centering
\includegraphics[width=0.42\textwidth]{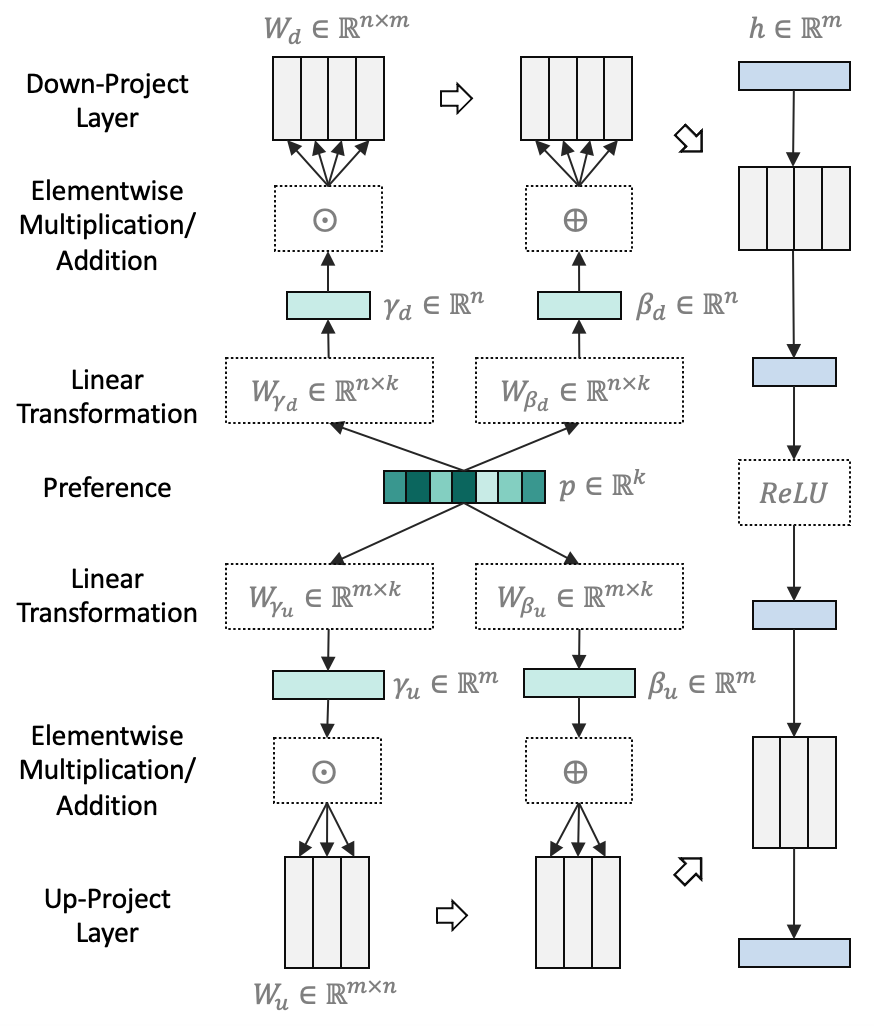}
\caption{Illustration of the operations of preference-aware adapters.}
\label{fg:adapter_operations}
\end{figure}

Given a pretrained base model and a preference vector $p$, we aim to design a module that can further introduce preference information into the summarizer. Meanwhile, the module should be compatible with the pretrained model without degrading performance. To achieve the requirements, we again leverage the adapter mentioned in Sec.~\ref{subsec:adapters}. The adapter is a bottlenecked feed-forward network consisting of a down-project layer (DP) and an up-project (UP) layer.
\begin{equation}
\begin{split}
    \text{DP}(h) & = W_{d} \cdot h, \\
    \text{UP}(\bar{h}) & = W_{u} \cdot \bar{h},
\end{split}
\end{equation}
where $h \in \mathbb{R}^m $ and $\bar{h} \in \mathbb{R}^n$ are linguistic features and $W_{d} \in \mathbb{R}^{n \times m}, W_{u} \in \mathbb{R}^{m \times n}$ are learnable parameters. A skip-connection from input to output is established, which can prevent noisy initialization from interfering with training at an early stage. The adapter (ADA) can thus be expressed as: 
\begin{equation}
    \text{ADA}(h) = \text{UP}(\text{ReLU}(\text{DP}(h))) + h.
\end{equation}
Based on the adapter, we propose leveraging the linear modulation~\cite{Perez_Strub_de_Vries_Dumoulin_Courville_2018} to extract the features in the context of the preferences. Specifically, the following transformations are conducted on parameters $W_d$ and $W_u$:
\begin{equation}
\begin{split}
    W_d^{'} & = (W_d \odot \gamma_d) \oplus \beta_d, \\  
    W_u^{'} & = (W_u \odot \gamma_u) \oplus \beta_u, \\  
\end{split}
\end{equation}
where $\odot$ and $\oplus$ are elementwise multiplication and addition operations, and the preference $p \in \mathbb{R}^k$ is incorporated in $\gamma_d$, $\beta_d$, $\gamma_u$ and $\beta_u$ as follows.
\begin{equation}
\begin{split}
    \gamma_d  & = W_{\gamma_d} \cdot p, \ \ \ \ \beta_d = W_{\beta_d} \cdot p, \\
    \gamma_u  & = W_{\gamma_u} \cdot p, \ \ \ \ \beta_u = W_{\beta_u} \cdot p, \\
\end{split}
\end{equation}
where $W_{\gamma_d}, W_{\beta_d} \in \mathbb{R}^{n \times k}, W_{\gamma_u}, W_{\beta_u} \in \mathbb{R}^{m \times k}$ are learnable parameters, and $\gamma_d, \beta_d \in \mathbb{R}^n, \gamma_u, \beta_u \in \mathbb{R}^m$ are used to perform transformation for the down-project and up-project layer. In other words, we project preference $p$ to create vectors for shifting and scaling the parameter distribution, and the summarizer can thus condition on the preference $p$. Since the preference contains much information about the required target, the summarizer would learn to create specific features with the modulated parameters according to the given preference. The preference-aware adapter (P-ADA) can be written as:
\begin{equation}
\begin{split}
    \text{P-DP}(h, p) & = W_{d}^{'} \cdot h, \\
    \text{P-UP}(\bar{h}, p) & = W_{u}^{'} \cdot \bar{h}, \\
\end{split}
\end{equation}
\begin{equation}
    \text{P-ADA}(h, p) = \text{P-UP}(\text{ReLU}(\text{P-DP}(h))) + h.
\end{equation}
The operations of the preference-aware adapter (P-ADA) are illustrated in Fig.~\ref{fg:adapter_operations}.

To further prevent the internal covariate shift~\cite{10.5555/3045118.3045167} arising from introduction of preference, we build a preference-aware layer normalization (P-LN) layer after the preference-aware adapters as:
\begin{equation}
    \text{P-LN}(h) = f(g' \odot \frac{h-\mu}{\sigma} + b'),
\end{equation}
where $g^{'} = W_{g} \odot g$, $b^{'} = W_{b} \odot b$, and $f$ is the activation function. The preference-aware adapter and normalization layer are used after each feed-forward layer in the decoder according to the learning method, which is elaborated in the next subsection. 

\subsection{Learning with Preferences}
Given the preference-aware summarizer and preferences describing the relations between articles and summaries, we propose two methods that leverage different sources of external knowledge for learning under low-resource scenarios. 

\subsubsection{Intra-Preference Learning (IPL)}
\label{subsec:self_supervised_transfer_learning}
\begin{figure}[t]
\centering
\includegraphics[width=0.40\textwidth]{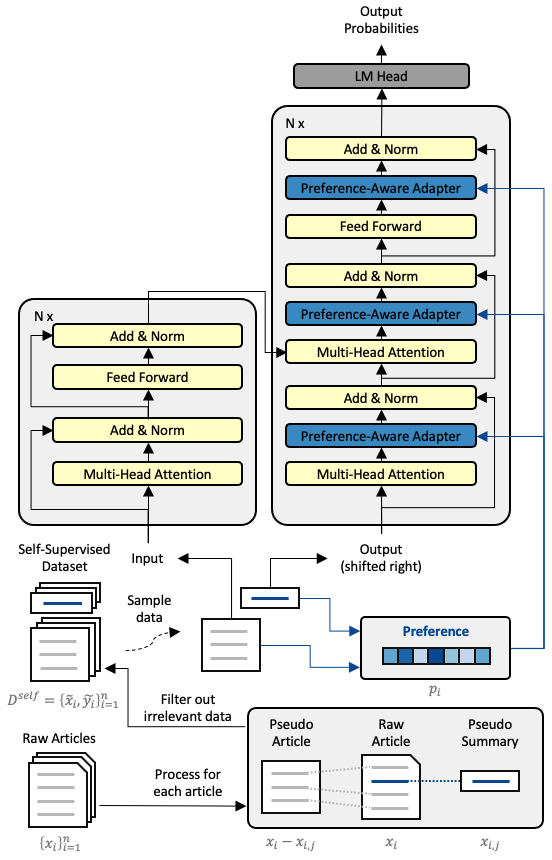}
\caption{Intra-preference learning framework.}
\label{fg:self_supervised_framework}
\end{figure}

In some summarization corpora, the summaries are automatically extracted from the raw articles with prior knowledge. For example, XSum~\cite{narayan-etal-2018-dont} uses the first sentences as the summaries according to the lead-bias, while CNN/DM~\cite{NIPS2015_afdec700} merges the highlight sentences before the main contents as the summaries. This inspires us that, since the sentences in the articles are essentially related to the article topic, 
the sentences could be treated as extreme summaries to provide a variety of preferences for the summarizer to learn. Based on this observation, we propose the Intra-Preferece Learning (IPL) to leverage such knowledge. Consider a partially-annotated target corpus $C^{tgt}=\{\{x_i\}_{i=1}^n, \{(x_i,y_i)\}_{i=n+1}^m\}$, where $n$ articles are without summaries, and $x_i=\{x_{i,1},x_{i,2},\cdots,x_{i,N_i}\}$ is the $i^{th}$ article that consists of $N_i$ complete sentences. We iteratively use each sentence as the \textit{pseudo summary}, and the remaining sentences as the \textit{pseudo article}, to create a self-supervised dataset $D^{self}$ as follows:
\begin{equation}
\begin{split}
    &\{(x_1-x_{1,1}, x_{1,1}), ( x_1-x_{1,2}, x_{1,2}), \cdots,( x_n-x_{n,N_n}, x_{n,N_n})\}\\
      & = \{ (\tilde{x}_1, \tilde{y}_1), (\tilde{x}_2, \tilde{y}_2), ...\}.
\end{split}
\end{equation}
We then compute the preferences for each example, and filter out those having zero ROUGE-1 scores, which are generally irrelevant to the article. During training, we insert the preference-aware adapters $\phi$ into the base model $\theta$ after each feed-forward layer in the decoder. The summarizer is then optimized to minimize the negative log-likelihood (NLL) on the $D^{self}$ as :
\begin{equation}
    -\log p(D^{self}|\theta, \phi) = -\sum_{(\tilde{x},\tilde{y}) \in D^{self}} \log p(\tilde{y}|\tilde{x},\theta, \phi).
    \label{eq:self_supervised_loss}
\end{equation}
After the self-supervised learning process, the summarizer is able to generate sentences according to the given preferences. However, the generated sentence is not like a summary under such a learning method. We thus use the few annotated examples $\{(x_i,y_i)\}_{i=n+1}^m$ to further finetune the summarizer to correct the generation formation by Eq.~\ref{eq:self_supervised_loss}. The architecture and the learning method are illustrated in Fig.~\ref{fg:self_supervised_framework}.

\begin{figure}[t]
\centering
\includegraphics[width=0.40\textwidth]{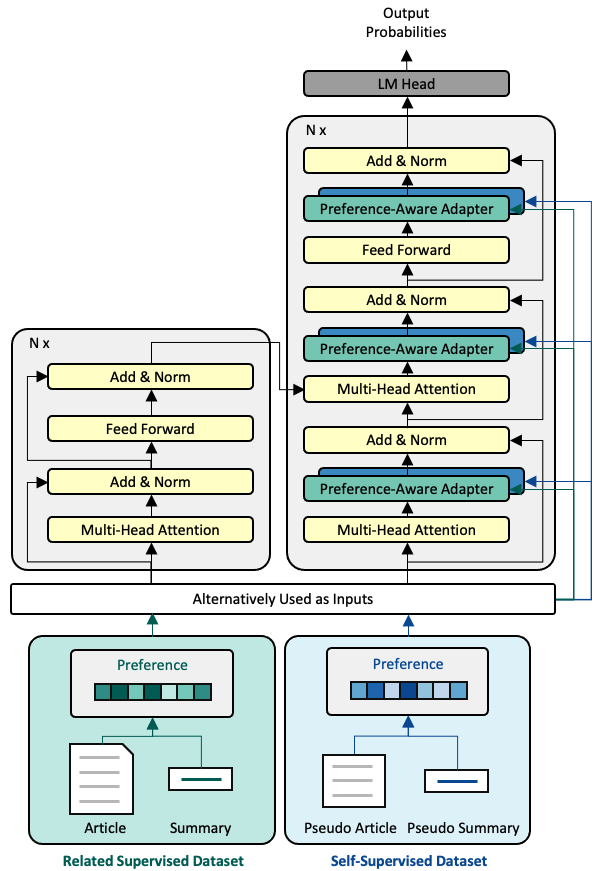}
\caption{Inter- and Intra-preference learning framework.}
\label{fg:meta_transfer_framework}
\end{figure}

\begin{algorithm}
\small
\caption{Inter- and Intra-Preference Learning}
\label{alg:spec_iipl}
\KwInput{Self-supervised dataset $D^{self}$, meta dataset $D^{meta}$, learning rate $\alpha, \beta, \gamma$}
\KwOutput{Base model $\theta$, preference-aware adapters $\phi^{self}, \phi^{meta}$}
Initialize $\theta$ with BART~\cite{lewis-etal-2020-bart}, randomly initialize $\phi^{self}$ and $\phi^{meta}$\;
\While{not converge}{
    \tcc{Intra-Phase}
    Randomly sample a batch of examples from $D^{self}$\;
    Evaluate loss and optimize $\theta, \phi^{self}$ by Eq.~\ref{eq:meta_self_sup_loss}\;
    \tcc{Inter-Phase}
    Sample a batch of tasks from $D^{meta}$ as $\mathcal{B}=\{D^{tr}_\tau, D^{te}_\tau\}$\;
    \For{each task-training set $D^{tr}_\tau$}{
        Initialize task-parameters $\psi$ with meta-parameters $\phi^{meta}$\;
        Evaluate loss by Eq.~\ref{eq:meta_inner_loss} and optimize $\psi$ with Eq.~\ref{eq:meta_inner_update}\;
        }
    Evaluate loss by Eq.~\ref{eq:meta_outer_loss} with $\{\psi_\tau\}$ and $\{D^{te}_\tau\}$ and optimize $\phi^{meta}$ by Eq.~\ref{eq:meta_outer_update}\;
}
\end{algorithm}

\subsubsection{Inter- and Intra-Preference Learning (IIPL)}
\label{subsec:iipl}
In parallel with the IPL that utilizes the intrinsic relations between articles and the subordinate sentences, it is also possible to leverage other annotated corpora for further improvement. These two sources essentially provide transfer knowledge from different perspectives. The self-supervised dataset helps models improve in-domain generation ability; other annotated corpora allow models to explore out-domain experiences to solve summarization tasks. To fully leverage the two sources with preference awareness, we propose the Inter- and Intra-Preference Learning (IIPL). Specifically, we perform meta learning with related annotated corpora in the context of features learned with the self-supervised dataset. Considering a partial annotated target corpus $C^{tgt}=\{\{x_i\}_{i=1}^N, \{(x_i,y_i)\}_{i=N+1}^M\}$, we use the method in Section \ref{subsec:self_supervised_transfer_learning} to create a self-supervised dataset $D^{self}$. Next, given a set of related annotated corpora, we need to choose suitable corpora to avoid negative transfer problems. We follow Sec.~\ref{subsec:meta_datasets} to use cosine similarity, ROUGE precision and article length for scoring corpora and then pick the top two as our source corpora $\{ C^{src}_j \}$. We then sample examples without replacement from $\{ C^{src}_j \}$ to create a collection of tasks $\{\mathcal{T}_\tau\}_{\tau=1}^M$ as the meta-dataset $D^{meta}$. Each task $\mathcal{T}_\tau$ contains a task-training set $D^{tr}_\tau=\{ (x^\tau_i,y^\tau_i) \}^K_{i=1}$ and a task-testing set $D^{te}_\tau=\{ ({x'}^{\tau}_i, {y'}^{\tau}_i) \}^K_{i=1}$. The number of tasks from different corpora is further balanced to avoid domain bias. To accommodate the two sources of knowledge in our summarizer, we insert two sets of adapters $\phi^{self}$ and $\phi^{meta}$ into the base model $\theta$ after each feed forward layer in the decoder. We then train $\phi^{self}$ and $\phi^{meta}$ with $D^{self}$ and $D^{meta}$ alternately at each step, which are referred as \textit{intra-phase} and \textit{inter-phase}, respectively. In the intra-phase, the summarizer is optimized to minimize the negative log-likelihood on the $D^{self}$ as follows:
\begin{equation}
    -\log p(D^{self}|\theta, \phi^{self}) = -\sum_{(\tilde{x},\tilde{y}) \in D^{self}} \log p(\tilde{y}|\tilde{x},\theta, \phi^{self}).
\label{eq:meta_self_sup_loss}
\end{equation}
During this phase, the base model shifts the features toward the in-domain data. In the inter-phase, it includes a bi-level optimization process. Given a batch of tasks $\mathcal{B}=\{\mathcal{T}_\tau | \tau \in B\}$. In the inner loop optimization, the task-parameters $\psi$ are initialized with meta-parameters $\phi^{meta}$ and optimized to minimize the following objective with task-training set $D^{tr}_\tau$ as: 
\begin{equation}
    -\log p(D^{tr}_\tau|\psi, \phi^{meta}) = -\sum_{(x,y) \in D^{tr}_\tau}\log p(y|x,\psi, \phi^{meta}).
\label{eq:meta_inner_loss}
\end{equation}
Through optimization, the task-parameters $\psi$ are adapted to the specific task $\tau$ as $\psi_\tau$. Assuming there is only one update step, it can be expressed as: 
\begin{equation}
    \psi_\tau \leftarrow \psi + \beta\nabla_\psi \log p(D^{tr}_\tau|\psi, \phi^{meta}),
\label{eq:meta_inner_update}
\end{equation}
where $\beta$ is the inner loop learning rate. For the outer loop optimization, meta-parameters $\phi^{meta}$ in another way minimize the testing loss after adaptation with all task-testing sets $D^{te}_\tau$ as follows:
\begin{equation}
    -\log p(\mathcal{B}|\phi^{meta})=-\sum_{\tau \in B} \log p(D^{te}_\tau|D^{tr}_\tau,\psi_\tau,\phi^{meta}).
\label{eq:meta_outer_loss}
\end{equation}
The meta-parameters $\phi^{meta}$ can then be updated by:
\begin{equation}
    \phi^{meta} \leftarrow \phi^{meta} + \alpha \nabla_{\phi^{meta}} \log p(\mathcal{B}|\phi^{meta}),
\label{eq:meta_outer_update}
\end{equation}
where $\alpha$ is the outer loop learning rate. During the inter-phase, the meta-parameters $\phi^{meta}$ will possess more generalization knowledge from different tasks. Since the two learning phases are of different purposes, the base model $\theta$ is fixed during the inter-phase. The dynamics of parameters caused by the intra-phase could in another way provide diverse observations for the meta-learner to improve generalizability under the in-domain context. As mentioned in Sec.~\ref{subsec:self_supervised_transfer_learning}, we use the few annotated examples $\{(x_i,y_i)\}_{i=N+1}^M$ to further finetune $\theta$ and $\phi^{meta}$ after the two-phase learning. The architecture and the learning method are illustrated in Fig.~\ref{fg:meta_transfer_framework}. The pseudo algorithm is shown in Alg.~\ref{alg:spec_iipl}.

\begin{figure}[t]
\centering
\includegraphics[width=0.40\textwidth]{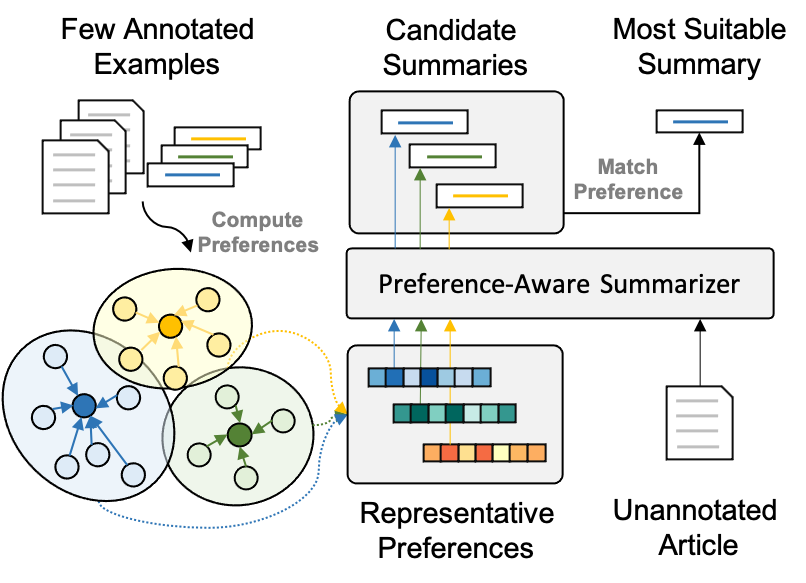}
\caption{Illustration of the preference-match decoding.}
\label{fg:pmd}
\end{figure}

\subsection{Preference-Match Decoding}
With the preference-aware summarizer, one could provide desired preferences to generate the corresponding summaries, which is beyond the typical singular preference generation process. However, in some cases, the preference may not be easy to define. For example, the proposed \textit{extractive diversity} and \textit{novel word ratio} actually conflict with each other. It is almost impossible to generate summaries that have high scores on these two properties at the same time. Such conflicts may not be easy to handle. Moreover, the preferences are essentially some scalars that could be difficult to obtain from observations. To solve this problem, we propose the Preference-Match Decoding (PMD), which utilizes only few annotated examples to automatically generate suitable candidate summaries for user selection. Specifically, we hypothesize that a corpus can contain few \textit{representative preferences}. The summaries generated given these representatives could be suitable under the data distribution. To find such representatives, we use K-Means clustering with the following objectives:
\begin{equation}
    \arg\min_\mathbf{S} \sum_{k=1}^K \sum_{p \in S_i} \| p-\mu_i \|^2,
\end{equation}
where $p$ is a preference, $\mathbf{S}=\{S_1, ..., S_K\}$ is the $K$ partitioned clusters, and $\mu_i$ is the mean of preferences in $S_i$. We use Elkan's algorithm~\cite{10.5555/3041838.3041857} to solve this optimization problem. Given an unannotated document $x$, we use the trained preference-aware summarizer (PAS) to generate summaries with the representatives $\{\mu_i\}$ as the candidate summaries. Users can thus select suitable sentences under the given preference distribution. To report the quantitative performance, we match the candidate summaries with the ground truth as:
\begin{equation}
    \hat{y}_i \sim \max_i(\text{ROUGE}(\text{PAS}(x, u_i)),y).
\end{equation}
Under this decoding process, we can ensure that the used preference is generalized from the annotated examples. In fact, we later show that preference-match decoding can outperform using ground truth preferences due to such a property in the experiment section. The operations of preference-match decoding are illustrated in Fig.~\ref{fg:pmd}.

\section{Experiments}
\subsection{Corpora}
We evaluate the proposed methods on several corpora, including Amazon Product Reviews~\cite{keung-etal-2020-multilingual}, Reddit-TIFU~\cite{kim-etal-2019-abstractive}, Movie Reviews~\cite{wang-ling-2016-neural}, SAMSum~\cite{gliwa-etal-2019-samsum}, SciTLDR~\cite{cachola-etal-2020-tldr}, and XSum~\cite{narayan-etal-2018-dont}. These corpora are chosen because they are from diverse domains, \textit{i.e.}, news (XSum), reviews (Amazon, Movie), online discussion (Reddit-TIFU), daily dialogue (SAMSum), and scientific papers (SciTLDR). Additionally, these corpora cover different levels of ROUGE performance, which can demonstrate the generality of the proposed methods on tasks of different difficulty levels. 

\subsection{Baselines}
\label{subsec:baseline}
The proposed \TASLPabbrev~aims to improve the generation qualities on the target corpus with few training examples. Therefore, we include self-supervised methods~\cite{lewis-etal-2020-bart,pmlr-v119-zhang20ae}, intermediate finetuning methods~\cite{fabbri-etal-2021-improving}, and meta learning methods (\AAAIabbrev) as the baselines for \TASLPabbrev. It is worth noting that we do not consider other summarization methods that introduce preferences for controlling the generations~\cite{fan-etal-2018-controllable,chan-etal-2021-controllable, jin-etal-2020-hooks, tan-etal-2020-summarizing, gao2020preference}. The reason is that our methods aim to evaluate with public datasets (human-written summaries), whereas the preferences considered in other methods are generally not applicable to public datasets. For example, \cite{jin-etal-2020-hooks} proposes to generate summaries in specific styles such as romance or humor, but the corresponding preference values on the public datasets are unknown. Moreover, \cite{tan-etal-2020-summarizing} proposes summarizing on arbitrary aspects relevant to the document with the aid of rich external knowledge sources such as ConceptNet and Wikipedia. However, the ground truths of the aspects for resulting summaries in public datasets are unavailable\footnote{The illustration of this claim is detailed in the Appendix.}.

\subsection{Implementation Details}
We implement \TASLPabbrev~and all baselines under the Pytorch framework. For fair comparisons, all Transformer architectures are reimplemented with the Huggingface Transformer library~\cite{wolf-etal-2020-transformers}. We use the large configuration for PEGASUS~\cite{pmlr-v119-zhang20ae} and the base configuration for BART~\cite{lewis-etal-2020-bart} to cover models with different capacities. For WikiTransfer, \AAAIabbrev, and \TASLPabbrev, we use the BART-base as the initialization. For the adapters in \AAAIabbrev~and preference-aware adapters in \TASLPabbrev, the hidden size is 64, and the parameters are initialized by a Gaussian distribution that has zero mean and a variance of 1e-4. For typical training before adaptation, a learning rate of 3e-5 and a batch size of 32 are used. The models are optimized with Adam~\cite{DBLP:journals/corr/KingmaB14} with early stopping for 9 epochs. The label smoothing method~\cite{7780677} is also applied with a factor of 0.1. For meta training before adaptation, the inner and outer learning rates are both set to 3e-5 with only one single gradient update in the inner loop. The base-learner is optimized with SGD, and the meta-learner is optimized with Adam for stability. For the low-resource adaptations, following previous works~\cite{fabbri-etal-2021-improving,pmlr-v119-zhang20ae}, we consider 10 and 100 annotated examples. For the parameter tuple (warm-up steps, total steps, learning rate), we use (25,100,3e-5) and (20,200,3e-5) for 10 and 100 examples, respectively, to finetune the \TASLPabbrev~and all baselines. For the preference-match decoding, the cluster number is set to 8. To construct meta-datasets for the \AAAIabbrev~and \TASLPabbrev, we use the method described in Sec.~\ref{subsec:meta_datasets} to rank 14 candidate corpora and choose the top two as the source corpora. 

\begin{table*}[t]
\scriptsize
\renewcommand\arraystretch{1.0}
\caption{Low-resource performance of \TASLPabbrev~compared with the previous SOTA on corpora in diverse domain.}\smallskip
\label{tb:low_resource_performance}
\centering
\begin{tabular}{p{0.08\textwidth}>{\centering}p{0.05\textwidth}ccc|ccc}
\toprule
\multirow{3}{*}{Corpus} & \multirow{3}{*}{\makecell{\# of \\ Annotated \\ Examples}} & \multicolumn{3}{c}{Without Human Annotations} & \multicolumn{3}{c}{With Human Annotations} \\
\cmidrule(l{1em}r{1em}){3-5}\cmidrule(l{1em}r{1em}){6-8}
& & BART~\cite{lewis-etal-2020-bart} & PEGASUS~\cite{pmlr-v119-zhang20ae} & \TASLPabbrev-IPL
  & WikiTransfer~\cite{fabbri-etal-2021-improving} & MTL-ABS & \TASLPabbrev-IIPL \\
& & $R_1$ / $R_2$ / $R_L$ & $R_1$ / $R_2$ / $R_L$ & $R_1$ / $R_2$ / $R_L$ 
  & $R_1$ / $R_2$ / $R_L$ & $R_1$ / $R_2$ / $R_L$ & $R_1$ / $R_2$ / $R_L$ \\
\midrule
\multirow{2}{*}{Amazon~\cite{keung-etal-2020-multilingual}} 
&  10 & 15.37/8.59/14.83 & 13.05/6.36/12.40 & \textbf{22.50/10.12/21.85} 
      & 15.02/8.26/14.48 & 13.94/6.02/13.70 & \textbf{24.01/11.14/23.30} \\
& 100 & 15.38/8.59/14.83 & 14.45/7.46/13.81 & \textbf{28.09/15.97/27.24} 
      & 15.95/7.96/15.56 & 14.74/6.82/14.42 & \textbf{27.47/15.23/26.73} \\
\midrule
\multirow{2}{*}{Reddit~\cite{kim-etal-2019-abstractive}} 
&  10 & 20.30/4.78/16.89 & 17.69/3.26/13.56 & \textbf{33.05/9.72/25.88} 
      & 21.59/5.52/17.97 & 21.38/5.47/17.64 & \textbf{33.44/10.28/25.86} \\
& 100 & 23.29/6.29/19.00 & 18.74/4.70/14.93 & \textbf{34.24/10.88/26.92} 
      & 22.34/6.12/18.53 & 22.21/6.13/18.45 & \textbf{34.02/10.70/26.61} \\
\midrule
\multirow{2}{*}{Movie~\cite{wang-ling-2016-neural}} 
&  10 & 23.80/7.66/19.16 & 18.65/4.34/14.36 & \textbf{30.67/11.01/23.88} 
      & 25.25/10.35/20.86 & 23.52/8.39/19.47 & \textbf{31.04/10.99/24.68} \\
& 100 & 26.05/9.06/20.82 & 20.16/5.85/15.82 & \textbf{31.29/11.67/23.94} 
      & 25.50/10.43/21.00 & 24.26/8.96/19.85 & \textbf{30.70/11.44/24.41} \\
\midrule
\multirow{2}{*}{SAMSum~\cite{gliwa-etal-2019-samsum}} 
& 10  & 34.90/12.00/31.43 & 27.91/7.40/23.63 & \textbf{46.06/20.90/40.34} 
      & 38.88/15.51/34.68 & 37.30/15.49/33.56 & \textbf{47.32/21.43/41.78} \\
& 100 & 44.26/18.59/40.10 & 41.82/18.69/37.66 & \textbf{51.94/24.75/46.97} 
      & 43.22/18.20/39.27 & 41.28/18.15/37.41 & \textbf{51.60/24.39/46.45} \\
\midrule
\multirow{2}{*}{SciTLDR~\cite{cachola-etal-2020-tldr}} 
& 10  & 33.45/13.46/27.89 & 25.70/7.75/19.47 & \textbf{41.81/19.60/34.88} 
      & 35.88/15.58/29.63 & 34.60/14.37/29.12 & \textbf{42.04/20.54/35.36} \\ 
& 100 & 38.08/17.84/32.00 & 32.40/13.77/26.19 & \textbf{44.85/23.43/37.91} 
      & 38.49/18.60/32.97 & 38.13/18.43/32.46 & \textbf{42.83/21.89/36.32} \\ 
\midrule
\multirow{2}{*}{XSum~\cite{narayan-etal-2018-dont}} 
& 10  & 26.00/7.47/20.31 & 19.04/3.15/13.45 & \textbf{32.74/10.90/24.86} 
      & 30.48/9.84/23.63 & 34.50/12.57/27.32 & \textbf{37.17/14.03/28.93} \\
& 100 & 32.43/10.96/25.44 & \textbf{38.87/17.60/31.38} & 35.69/12.88/27.25 
      & 32.82/10.85/25.58 & 34.70/12.63/27.47 & \textbf{37.48/14.35/29.12} \\
\bottomrule
\end{tabular}
\end{table*}

\subsection{Evaluation Plan}
\label{subsec:evaluation_plan}
In the following, we first compare \TASLPabbrev~with baselines in Sec.~\ref{subsec:main_result}. Next, we demonstrate that preference decomposition can enable successful transferring from self-supervised datasets in Sec.~\ref{subsec:preference_help_transfer} and visualize the preferences in Sec.~\ref{subsec:preference_visualization}. Then, to further study the design dimensions of the preference-match decoding, we study performance with different settings in Sec.~\ref{subsec:pdm_design_dimension}. Furthermore, we investigate the difference between using gold and representative preferences in Sec.~\ref{subsec:gold_representative_comparison}. Finally, we present generation cases to demonstrate the ability of \TASLPabbrev~to control the preferences in Sec.~\ref{subsec:generation_case}. Additional experiments, \textit{e.g.}, meta-transferability for Sec.~\ref{subsec:meta_datasets} and multi-target summarization, are shown in the Appendix.

\subsection{Main Results}
\label{subsec:main_result}
Table~\ref{tb:low_resource_performance} shows the performance of \TASLPabbrev~and all baselines on 6 corpora under low-resource settings. The results are grouped according to whether human annotations are used for ease of comparison. Specifically, the intra-preference learning (\TASLPabbrev-IPL) is grouped with the BART and PEGASUS, and the inter- and intra-preference learning (\TASLPabbrev-IIPL) is grouped with the WikiTransfer and \AAAIabbrev. The results demonstrate that, without human annotations, \TASLPabbrev-IPL outperforms the BART and PEGASUS on all corpora with 36.83\%/55.10\%/33.50\% and 29.45\%/37.44\%/27.13\% on ROUGE-1/2/L under 10- and 100-example settings, respectively. The improvements could be contributed jointly from the content-preference decomposition and the utilization of unannotated target documents. Specifically, the target documents provide knowledge that is distinct for the application, and the decomposition helps models transfer the knowledge without being constrained by the text properties in the document sentences that are drastically different from summaries. The results also manifest that the preference-match decoding can generalize well with only few preferences from the target corpus. It is worth noting that there is one exception for performance improvements, \textit{i.e.}, the PEGASUS slightly outperforms \TASLPabbrev-IPL under the 100-example setting on the XSum corpus. This is probably because the PEGASUS is pretrained with C4~\cite{JMLR:v21:20-074} and HugeNews~\cite{pmlr-v119-zhang20ae}, which consists of 350M web pages and 1.5B news articles, respectively. As the HugeNews corpus is in the same domain as the XSum corpus, pretraining may significantly contribute to the performance of the PEGASUS. Furthermore, the study of \cite{pmlr-v119-zhang20ae} also shows that C4 and XSum have a significant amount of overlap (approximately 15\% to 20\%), which provides a large amount of self-supervised signals extremely similar to the XSum corpus. However, \TASLPabbrev~leverages only around 200K unannotated documents to achieve comparable performance and still surpasses the PEGASUS under the 10-example setting. It is also worth noting that we can further improve \TASLPabbrev-IPL to a ROUGE-1 score of 38.33, which is at the same level as the PEGASUS, by increasing the cluster number in preference-match decoding from 8 to 16 (later described in Sec.~\ref{subsec:pdm_design_dimension}).

Furthermore, with the combination of meta learning designs from \AAAIabbrev, \TASLPabbrev-IIPL could improve over all baselines including \AAAIabbrev~with 30.11\%/33.95\%/27.51\% and 26.74\%/31.14\%/24.48\% on ROUGE-1/2/L under 10- and 100-example settings, respectively. Compared to the \AAAIabbrev, which leverages existing corpora and freezes the base model throughout learning, \TASLPabbrev-IIPL~learns the intra-phase adapters together with the base model on the self-supervised datasets, and the inter-phase adapters can therefore learn from diverse corpora in the context of in-domain features. \TASLPabbrev-IIPL makes further improvements over \TASLPabbrev-IPL in 10-example settings, showing that our methods successfully coordinate two sources of knowledge to improve performance. For 100-example settings, the advantages from related annotated corpora diminish for most corpora. This is probably because of the task size used during the inter-phase. Each task-training and task-testing set in our setting contains 8 examples, which can simulate the generalization behavior well under 10-example settings but not necessarily for 100-example settings. It is worth noting that performing bi-level optimization on such a scale is impractical due to the significant memory overhead. We left this issue as future work for further study.

\subsection{Preference Decomposition Helps Transferring}
\label{subsec:preference_help_transfer}
A predominant challenge in both transfer and meta learning is the \textit{negative-transfer problem}. It is a phenomenon that transferring knowledge from sources can have a negative impact on the target learner~\cite{Wang_2019_CVPR}. In \TASLPabbrev, we utilize document sentences as pseudo summaries, and this could generally result in the negative-transfer problem. To verify that \TASLPabbrev~can overcome this problem through the introduction of preferences, we use the self-supervised datasets proposed in Sec.~\ref{subsec:self_supervised_transfer_learning} to learn a Transformer model initialized with the BART~\cite{lewis-etal-2020-bart} and name it BART-SS. The performance comparisons between BART-SS and \TASLPabbrev-IPL are shown in Table~\ref{tb:preference_help_transfer}. The results manifest that \TASLPabbrev~surpasses the BART-SS on all corpora, which could be the benefit of decomposing contents and preferences. This claim could also be demonstrated by comparing the BART-SS with the BART baseline in Table~\ref{tb:low_resource_performance}, where the difference between these two baselines is that BART-SS is first finetuned with self-supervised datasets and then with the target corpus, while the one in Table~\ref{tb:low_resource_performance} is only finetuned on the target corpus. From the comparisons, we could observe that direct finetuning with the self-supervised datasets could indeed cause a negative-transfer problem and degrade the performance. This is particularly severe for corpora such as Amazon and Reddit, in which the style of document sentences is totally different from summaries. \TASLPabbrev~successfully prevents this problem by implicitly decomposing the contextual understating and writing style and can thus utilize the shared knowledge between source and target corpora for further improvements.

\begin{table}[h]
\scriptsize
\renewcommand\arraystretch{1.0}
\caption{The introduction of preferences can effectively prevent the negative-transfer problem and provide further improvements.}\smallskip
\label{tb:preference_help_transfer}
\centering
\begin{tabular}{lccc}
\toprule
\multirow{2}{*}{\makecell{Corpus}} & \multirow{2}{*}{\makecell{Annotated \\ Examples}} & BART-SS & \TASLPabbrev-IPL  \\
& & $R_1$ / $R_2$ / $R_L$ & $R_1$ / $R_2$ / $R_L$ \\
\midrule
\multirow{2}{*}{Amazon~\cite{keung-etal-2020-multilingual}} 
& 10  & 6.98/1.68/6.54 & \textbf{22.50/10.12/21.85} \\
& 100 & 12.51/5.78/12.19 & \textbf{28.09/15.97/27.24} \\
\midrule
\multirow{2}{*}{Reddit~\cite{kim-etal-2019-abstractive}} 
& 10  & 18.33/3.75/15.22 & \textbf{33.05/9.72/25.88}  \\
& 100 & 22.18/5.51/18.13 & \textbf{34.24/10.88/26.92} \\
\midrule
\multirow{2}{*}{Movie~\cite{wang-ling-2016-neural}} 
& 10  & 25.64/9.34/20.99 & \textbf{30.67/11.01/23.88} \\
& 100 & 25.36/8.81/20.14 & \textbf{31.29/11.67/23.94} \\
\midrule
\multirow{2}{*}{SAMSum~\cite{gliwa-etal-2019-samsum}} 
& 10  & 34.13/12.74/30.51 & \textbf{46.06/20.90/40.34} \\
& 100 & 44.63/19.13/40.28 & \textbf{51.94/24.75/46.97} \\
\midrule
\multirow{2}{*}{SciTLDR~\cite{cachola-etal-2020-tldr}} 
& 10  & 35.11/14.83/28.73 & \textbf{41.81/19.60/34.88} \\
& 100 & 37.91/18.01/31.93 & \textbf{44.85/23.43/37.91} \\
\midrule
\multirow{2}{*}{XSum~\cite{narayan-etal-2018-dont}} 
& 10  & 29.05/9.06/22.32  & \textbf{32.74/10.90/24.86}  \\
& 100 & 32.35/11.25/25.12 & \textbf{35.69/12.88/27.25}  \\
\bottomrule
\end{tabular}
\end{table}

\subsection{Visualization of Preferences}
\label{subsec:preference_visualization}
\begin{figure}[h]
\centering
\includegraphics[width=0.45\textwidth]{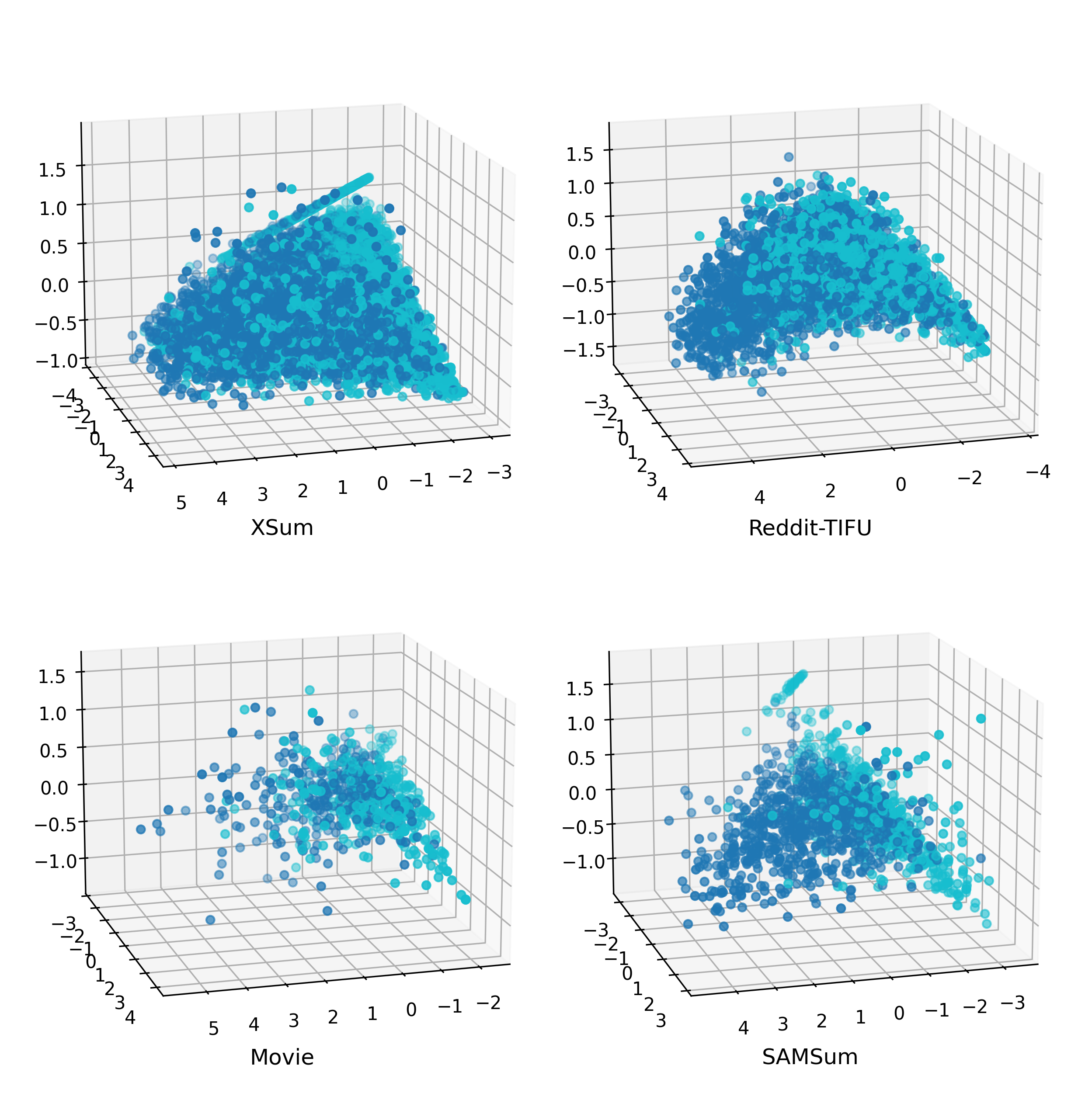}
\caption{Visualization of preferences. The preferences are projected into 3D dimensions through the principal components analysis (PCA). The preferences of self-supervised datasets are colored in aqua, and those of summaries are colored in blue. Best viewed in color.}
\label{fg:preferences_visualization}
\end{figure}

For further studies, we visualize the preferences for four of the target corpus and the corresponding self-supervised datasets in Fig.~\ref{fg:preferences_visualization}. From the results, the preferences of the self-supervised dataset can generally cover those of summaries for the XSum corpus. However, for the Reddit and Movie corpora, the overlaps are relatively low, and for the SAMSum corpus, the characteristics of document sentences are generally dissimilar to the summaries. This observation meets the results in Sec.~\ref{subsec:preference_help_transfer} that preferences become especially important to prevent the negative-transfer problem on those corpora.

\subsection{Design Dimensions of Preference-Match Decoding}
\label{subsec:pdm_design_dimension}

\begin{table*}[t]
\scriptsize
\renewcommand\arraystretch{1.0}
\caption{Performance of \TASLPabbrev~with different number of \textbf{available preferences from target corpora}. $\dagger$ indicates there are additional available target preferences under that low-resource setting.}\smallskip
\label{tb:pmd_available_num}
\centering
\begin{tabular}{ccccccc}
\toprule
\multirow{2}{*}{\makecell{\# of Available \\ Preferences}} &
    Amazon~\cite{keung-etal-2020-multilingual} &
    Reddit~\cite{kim-etal-2019-abstractive} &
    Movie~\cite{wang-ling-2016-neural} &
    SAMSum~\cite{gliwa-etal-2019-samsum} &
    SciTLDR~\cite{cachola-etal-2020-tldr} &
    XSum~\cite{narayan-etal-2018-dont} \\
&   $R_1$ / $R_2$ / $R_L$ &
    $R_1$ / $R_2$ / $R_L$ &
    $R_1$ / $R_2$ / $R_L$ &
    $R_1$ / $R_2$ / $R_L$ &
    $R_1$ / $R_2$ / $R_L$ &
    $R_1$ / $R_2$ / $R_L$ \\
\midrule
\multicolumn{7}{c}{\textit{10-example setting}} \\
\midrule
10 &
    22.50/10.12/21.85 &
    33.05/9.72/25.88 &
    30.67/11.01/23.88 &
    46.06/20.90/40.34 &
    41.81/19.60/34.88 &
    32.74/10.90/24.86 \\
\midrule
100 $\dagger$  &
    24.57/12.53/23.85 &
    32.80/9.44/25.86 &
    30.92/11.28/24.02 &
    46.09/20.71/40.44 &
    42.31/20.93/35.73 &
    32.23/10.65/24.55 \\
\midrule
1000 $\dagger$ &
    22.79/11.43/22.01 &
    33.00/9.55/25.76 &
    30.45/10.74/23.45 &
    46.03/20.84/40.51 &
    42.02/20.46/35.28 &
    32.32/10.50/24.44 \\
\midrule
\multicolumn{7}{c}{\textit{100-example setting}} \\
\midrule
10 &
    27.12/14.19/26.51 &
    34.48/11.12/26.96 &
    32.01/11.73/24.63 &
    52.34/24.87/47.16 &
    44.13/22.14/37.09 &
    36.20/13.16/27.59 \\
\midrule
100 &
    28.09/15.97/27.24 &
    34.24/10.88/26.92 &
    31.29/11.67/23.94 &
    51.94/24.75/46.97 &
    44.85/23.43/37.91 &
    35.69/12.88/27.25 \\
\midrule
1000 $\dagger$ &
    26.48/15.11/25.62 &
    34.48/10.90/27.04 &
    30.81/11.28/23.87 &
    52.23/24.72/47.04 &
    44.56/23.15/37.55 &
    35.66/12.82/27.00 \\
\bottomrule
\end{tabular}
\end{table*}

\begin{table*}[t]
\scriptsize
\renewcommand\arraystretch{1.0}
\caption{Performance of \TASLPabbrev~with different number of \textbf{representative preferences estimated from target corpora}.}\smallskip
\label{tb:pmd_representative_num}
\centering
\begin{tabular}{ccccccc}
\toprule
\multirow{2}{*}{\makecell{\# of Representative \\ Preferences}} &
    Amazon~\cite{keung-etal-2020-multilingual} &
    Reddit~\cite{kim-etal-2019-abstractive} &
    Movie~\cite{wang-ling-2016-neural} &
    SAMSum~\cite{gliwa-etal-2019-samsum} &
    SciTLDR~\cite{cachola-etal-2020-tldr} &
    XSum~\cite{narayan-etal-2018-dont} \\
&   $R_1$ / $R_2$ / $R_L$ &
    $R_1$ / $R_2$ / $R_L$ &
    $R_1$ / $R_2$ / $R_L$ &
    $R_1$ / $R_2$ / $R_L$ &
    $R_1$ / $R_2$ / $R_L$ &
    $R_1$ / $R_2$ / $R_L$ \\
\midrule
\multicolumn{7}{c}{\textit{10-example setting}} \\
\midrule
1 &
    11.33/3.35/10.92 &
    26.31/6.40/21.17 &
    24.89/8.00/19.54 &
    37.80/14.61/33.23 &
    34.93/14.87/29.33 &
    25.11/6.83/18.97 \\
\midrule
2 &
    15.87/7.04/15.34 &
    28.07/7.37/22.55 &
    27.21/8.94/21.42 &
    42.07/17.99/37.23 &
    38.57/17.42/32.38 &
    27.93/8.35/21.34 \\
\midrule
4 &
    20.40/9.48/19.91 &
    31.06/8.65/24.50 &
    29.27/10.18/22.74 &
    43.86/19.39/38.36 &
    39.79/18.28/33.35 &
    30.76/9.63/23.22 \\
\midrule
8 &
    22.79/11.43/22.01 &
    33.00/9.55/25.76 &
    30.45/10.74/23.45 &
    46.03/20.84/40.51 &
    42.02/20.46/35.28 &
    32.32/10.50/24.44 \\
\midrule
\multicolumn{7}{c}{\textit{100-example setting}} \\
\midrule
1 &
    14.21/5.87/13.72 &
    27.46/7.34/21.69 &
    24.48/8.01/19.31 &
    43.98/18.25/39.44 &
    37.77/17.66/31.84 &
    28.05/8.30/21.05 \\
\midrule
2 &
    19.12/9.76/18.47 &
    30.61/8.79/23.98 &
    28.49/10.50/22.26 &
    47.23/20.74/42.52 &
    41.42/20.90/35.02 &
    30.83/10.27/23.73 \\
\midrule
4 &
    22.68/12.67/21.98 &
    32.62/9.81/25.48 &
    30.32/11.34/23.33 &
    49.80/22.98/45.14 &
    43.28/22.39/36.58 &
    33.63/11.67/25.77 \\
\midrule
8 &
    26.48/15.11/25.62 &
    34.48/10.90/27.04 &
    30.81/11.28/23.87 &
    52.23/24.72/47.04 &
    44.56/23.15/37.55 &
    35.66/12.82/27.00 \\
\midrule
16 &
    31.68/18.02/30.79 &
    36.06/11.80/28.20 &
    32.89/12.25/25.19 &
    54.02/26.30/49.00 &
    46.04/24.10/38.88 &
    38.33/14.66/29.30 \\
\bottomrule
\end{tabular}
\end{table*}

The preference-match decoding will estimate the \textit{representative preferences} from the \textit{available preferences} of the target corpus, in the following, we further study the performance with different number of available and representative preferences in Table~\ref{tb:pmd_available_num} and~\ref{tb:pmd_representative_num}, respectively. 

\subsubsection{Different Number of Available Preferences}
Table~\ref{tb:pmd_available_num} shows the performance with 10/100/1000 available preferences under 10- and 100-example settings. It is worth noting that we also include cases where the number of available preferences is greater than the number of available annotated examples, \textit{e.g.}, 10 examples with 100 available preferences. These cases are considered to study how additional preference supervision would affect performance. From the results of the 10-example setting, we could observe that the performance does not improve as the number of available preferences increases, \textit{i.e.} to 100 and 1000. This suggests that the preference-match decoding can well estimate and leverage the representative preferences even with only few examples. Similar observations could also be obtained under the 100-example setting, where using only 10 target preferences can achieve a similar performance with the 100 and 1000 counterparts.

\subsubsection{Different Number of Representative Preferences}
Table V shows the performance with 1/2/4/8/16 representative preferences estimated under 10- and 100-example settings. The results show that the performance can be improved as we increase the number of representative preferences under both settings, which meets our expectations since the target preference distribution could be better explored. Meanwhile, by reducing the number of representative preferences to 1, the performance would therefore be degraded to the same level as the BART-base results in Table~\ref{tb:low_resource_performance}. Comparing 10- and 100-example settings, the latter has a larger room for performance improvement due to the higher data diversity. It is worth noting that the performance benefits from a higher number of representative preferences, but it also aggravates the computation costs. To demonstrate such a tradeoff, we follow a previous work~\cite{pmlr-v80-kaiser18a} to report the decoding time for data averaged over the whole testing set with different representative numbers in Table~\ref{tb:decoding_time}. The decoding is conducted on one NVIDIA RTX 3080 GPU, while the batch size is set to 16. The result suggests that the decoding time increases approximately linearly with the representative numbers. Therefore, in other experiments, we estimate 8 representatives since the performance appears to saturate from 8 to 16 clusters.

\begin{table}[h]
\scriptsize
\renewcommand\arraystretch{1.0}
\caption{The average decoding time per data with different number of representative preferences.}\smallskip
\label{tb:decoding_time}
\centering
\begin{tabular}{lc}
\toprule
\multirow{2}{*}{Corpus} & \# of Representative Preferences \\
& 1 / 2 / 4 / 8 / 16 \\
\midrule
Amazon~\cite{keung-etal-2020-multilingual} &
72 / 103 / 213 / 545 / 1147 ms \\
\midrule
Reddit~\cite{kim-etal-2019-abstractive} &
204 / 325 / 775 / 1603 / 2941 ms \\ 
\midrule
Movie~\cite{wang-ling-2016-neural} &
214 / 372 / 717 / 1307 / 2695 ms \\ 
\midrule
SAMSum~\cite{gliwa-etal-2019-samsum} &
193 / 362 / 715 / 1290 / 2740 ms \\ 
\midrule
SciTLDR~\cite{cachola-etal-2020-tldr} &
186 / 333 / 651 / 1279 / 2611 ms \\ 
\midrule
XSum~\cite{narayan-etal-2018-dont} &
239 / 396 / 797 / 1546 / 3096 ms \\
\bottomrule
\end{tabular}
\end{table}

\subsection{Comparisons between Gold and Representative Preferences}
\label{subsec:gold_representative_comparison}

\begin{table}[h]
\scriptsize
\renewcommand\arraystretch{1.0}
\caption{Performance given gold and representative preferences.}\smallskip
\label{tb:gold_represent_performance}
\centering
\begin{tabular}{lccc}
\toprule
\multirow{2}{*}{Corpus} 
& \multirow{2}{*}{\makecell{Annotated \\ Examples}} & Gold & Representative \\
& & $R_1$ / $R_2$ / $R_L$ & $R_1$ / $R_2$ / $R_L$  \\
\midrule
\multirow{2}{*}{Amazon~\cite{keung-etal-2020-multilingual}}
& 10 & 17.56/9.07/17.16 & \textbf{22.79/11.43/22.01}  \\
& 100 & 22.82/13.58/22.36 & \textbf{26.48/15.11/25.62}  \\
\midrule
\multirow{2}{*}{Reddit~\cite{kim-etal-2019-abstractive}}
& 10 & 27.94/6.99/22.12 & \textbf{33.00/9.55/23.80}  \\
& 100 & 29.77/8.37/23.43 & \textbf{34.48/10.90/27.04}  \\
\midrule
\multirow{2}{*}{Movie~\cite{wang-ling-2016-neural}}
& 10 & 25.58/8.85/20.25 & \textbf{30.45/10.74/23.45}  \\
& 100 & 25.07/8.66/19.82 & \textbf{30.81/11.28/23.87}  \\
\midrule
\multirow{2}{*}{SAMSum~\cite{gliwa-etal-2019-samsum}} 
& 10 & 40.52/15.68/35.68 & \textbf{46.03/20.84/40.51}  \\
& 100 & 46.80/19.86/42.14 & \textbf{52.23/24.72/47.04}  \\
\midrule
\multirow{2}{*}{SciTLDR~\cite{cachola-etal-2020-tldr}} 
& 10 & 35.59/15.56/29.92 & \textbf{42.02/20.46/35.28}  \\
& 100 & 38.25/18.56/32.51 & \textbf{44.56/23.15/37.55}  \\
\midrule
\multirow{2}{*}{XSum~\cite{narayan-etal-2018-dont}} 
& 10 & 27.96/8.02/21.21 & \textbf{32.32/10.50/24.44}  \\
& 100 & 31.63/10.13/23.86 & \textbf{35.66/12.82/27.00}  \\
\bottomrule
\end{tabular}
\end{table}

\begin{table*}[t]
\scriptsize
\renewcommand\arraystretch{1.0}
\caption{Generation cases using gold and representative preferences. Preferences are shown in the order of ROUGE-1, ROUGE-2, ROUGE-L, compression ratio, extractive coverage, extractive density, and novel word ratio.}\smallskip
\label{tb:gold_represent_examples}
\centering
\begin{tabular}{p{0.97\textwidth}}
\toprule
\multicolumn{1}{c}{Reddit} \\
\toprule
\textbf{Ground Truth:} Never try to dance with a bench. there go my pecs. \\ 
\textbf{Gold Preference:} 0.10/0.01/0.05/0.05/0.64/0.82/0.27 \\
\midrule
\textbf{Generated Summary Given Gold Preference:} I sprained my pectoralis minor and ended up in the hospital. \\
\textbf{ROUGE-1/2/L:} 9.09/0.0/9.09 \\
\textbf{Resulted Preference:} 0.11/0.04/0.05/0.05/0.72/2.0/0.27 \\
\midrule
\textbf{Generated Summary Given Representative Preference:} Tried to create a dance with a bench, ended up spraining my pectoralis. \\
\textbf{ROUGE-1/2/L:} 49.99/27.27/49.99 \\
\textbf{Resulted Preference:} 0.13/0.03/0.08/0.06/0.77/1.23/0.23 \\
\toprule
\multicolumn{1}{c}{XSum} \\
\toprule
\textbf{Ground truth:} New guidance has been offered to landowners to help them avoid clashes with dog owners. \\ 
\textbf{Gold Preference:} 0.08/0.01/0.04/0.04/0.67/0.93/0.33 \\
\midrule
\textbf{Generated Summary Given Gold Preference:} Dog walkers have been urged to reduce the number of incidents caused by "irresponsible" owners. \\ 
\textbf{ROUGE-1/2/L:} 26.67/0.0/20.00 \\
\textbf{Resulted Preference:} 0.08/0.01/0.03/0.04/0.67/1.2/0.33 \\
\midrule
\textbf{Generated Summary Given Representative Preference:} NFU Scotland has issued guidance to reduce the number of dog-walking incidents. \\
\textbf{ROUGE-1/2/L:} 28.57/0.0/21.43 \\
\textbf{Resulted Preference:} 0.07/0.02/0.04/0.03/0.75/1.42/0.25 \\
\bottomrule
\end{tabular}
\end{table*}

\begin{figure}[h]
\centering
\subfloat[Preference recover error\label{fg:preference_recover_error}]{%
\includegraphics[width=0.48\linewidth]{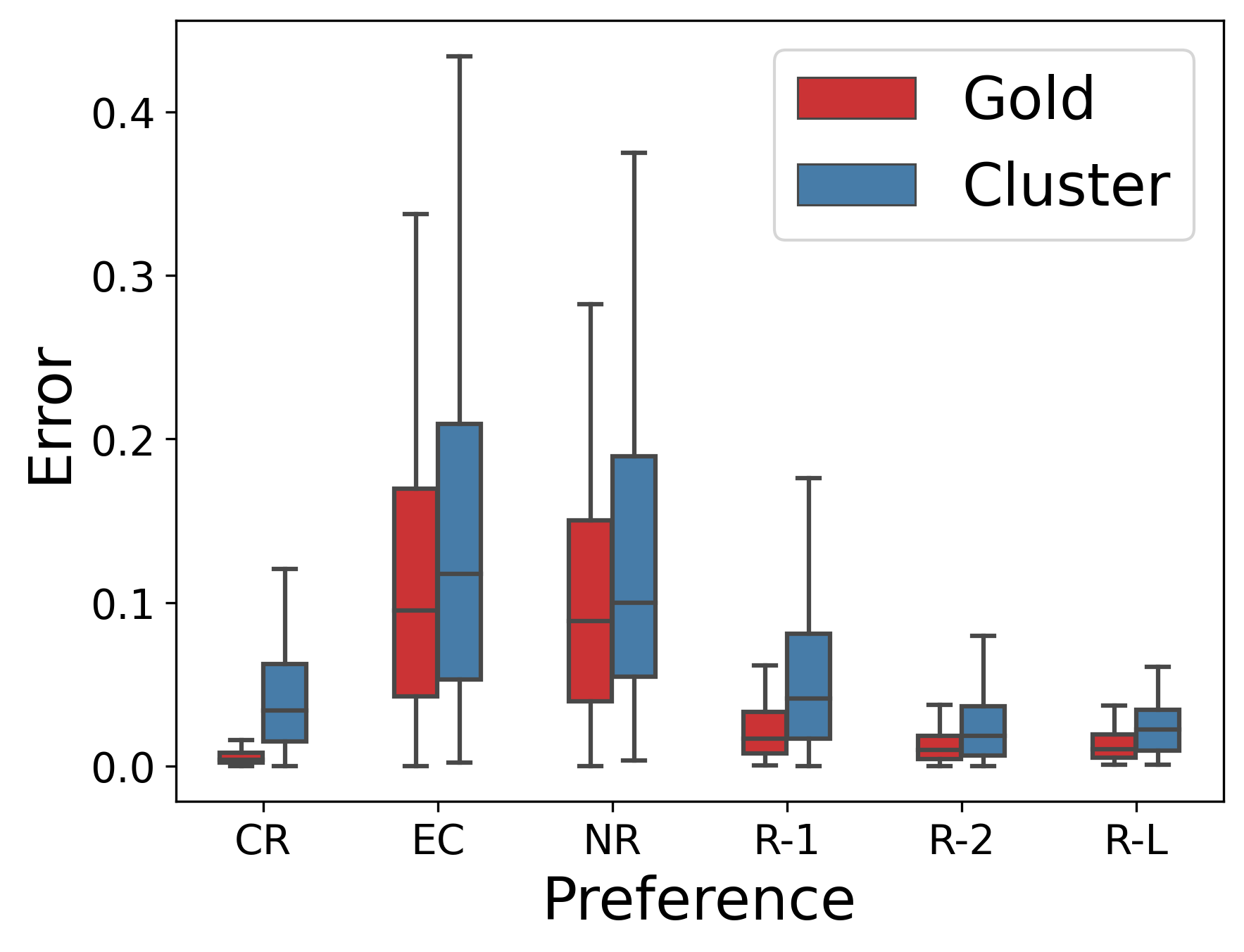}}
\hspace{-3pt}
\subfloat[Token-level loss\label{fg:token_loss}]{%
\includegraphics[width=0.48\linewidth]{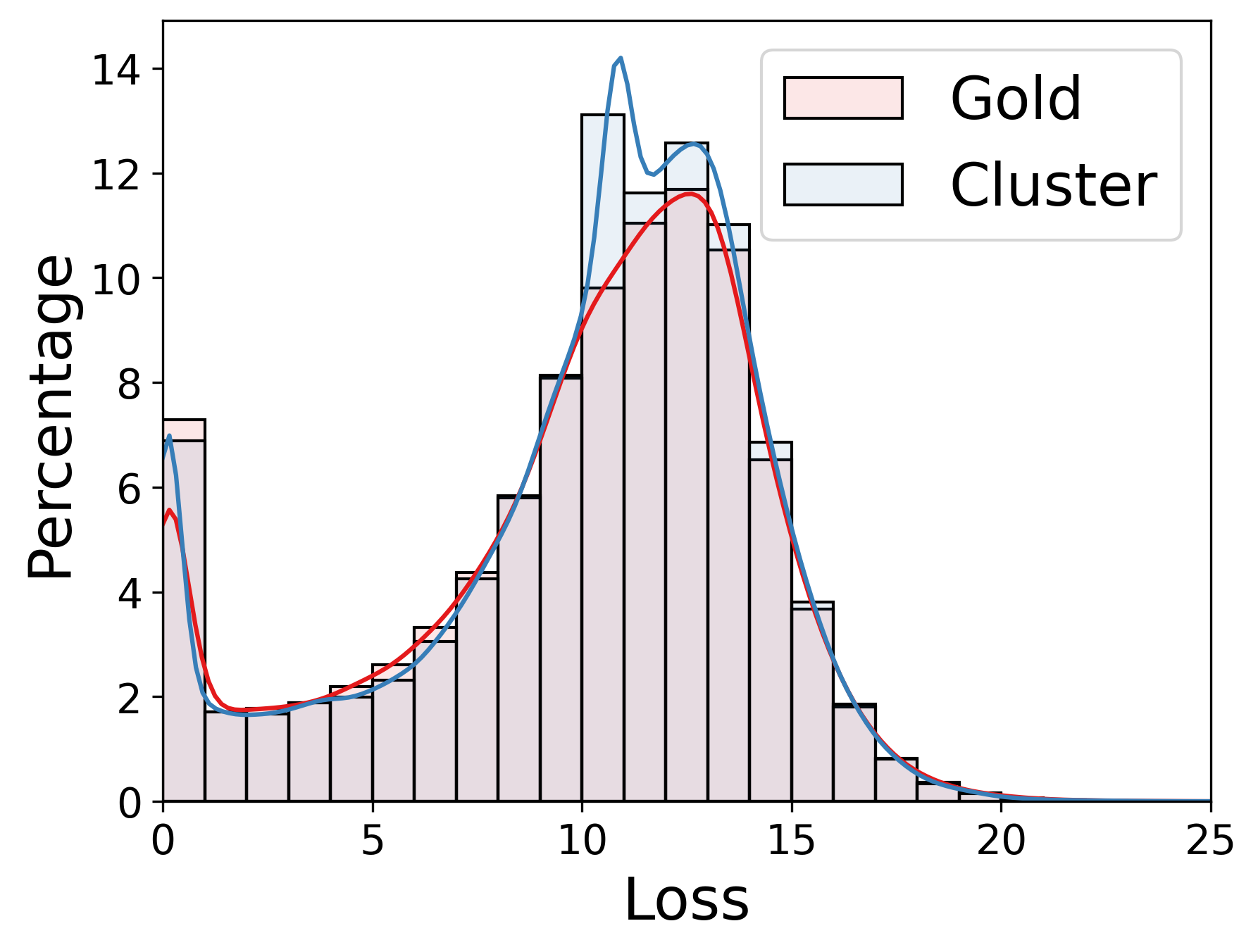}}
\caption{(a) Preference recover error given gold or representative (cluster) preferences on XSum corpus. The box plot shows error statistics for compress ratio (CR), extractive coverage (EC), novel word ratio (NR), ROUGE-1 (R-1), ROUGE-2 (R-2) and ROUGE-L (R-L). (b) Token-level negative log-likelihood given gold or representative preferences on XSum corpus.}
\end{figure}

The preference-match decoding creates summary candidates during inference according to the representative preferences estimated from few target examples. In the following, we further investigate the performance on the acquisition of gold preferences during inference. In other words, we directly provide gold preferences to the preference-aware summarizer, and the comparisons with \TASLPabbrev-IPL are shown in Table~\ref{tb:gold_represent_performance}. Intuitively, using gold preferences should perform better since the representative preferences are estimations. However, the results show that preference-match decoding achieves a higher performance. To study this phenomenon, we first compute the \textit{preference recover error}, which is the L1 distance between resulted and gold preferences, on the XSum corpus in Fig.~\ref{fg:preference_recover_error}. Also, we demonstrate two generation cases on the Reddit and XSum corpora given gold and representative preferences in Table~\ref{tb:gold_represent_examples}. We could observe from Fig.~\ref{fg:preference_recover_error} that providing gold preferences indeed makes resulted preferences closer to the ground truths since the average errors are smaller. However, the ROUGE scores in Table~\ref{tb:gold_represent_examples} manifest that reconstructing the gold preferences better does not necessarily imply that the summary qualities are better. This is probably because the representative preferences are more likely to locate in the dense regions of latent space and thus more likely to be explored than the gold preferences, leading to better generalization. This explanation is connected to a previous work studying the problem or text variational autoencoders (VAEs), where similar observations can also be found~\cite{10.5555/3524938.3525914}. Specifically, the previous work proposes the \textit{Latent Vacancy Hypothesis} to explain the generalization difficulty for text generation with VAEs. The main idea is that the aggregated posterior of text-VAEs tend to have vacant/low-density regions due to the discrete nature of text data. As such, it is difficult for the decoder to generalize to those regions. In our case, the preferences connected to the text may not be continuous in the latent space also. Once the downstream preferences fall in the vacancy of the space, the model could not generalize well. 

\begin{figure}[h]
\centering
\includegraphics[width=0.45\textwidth]{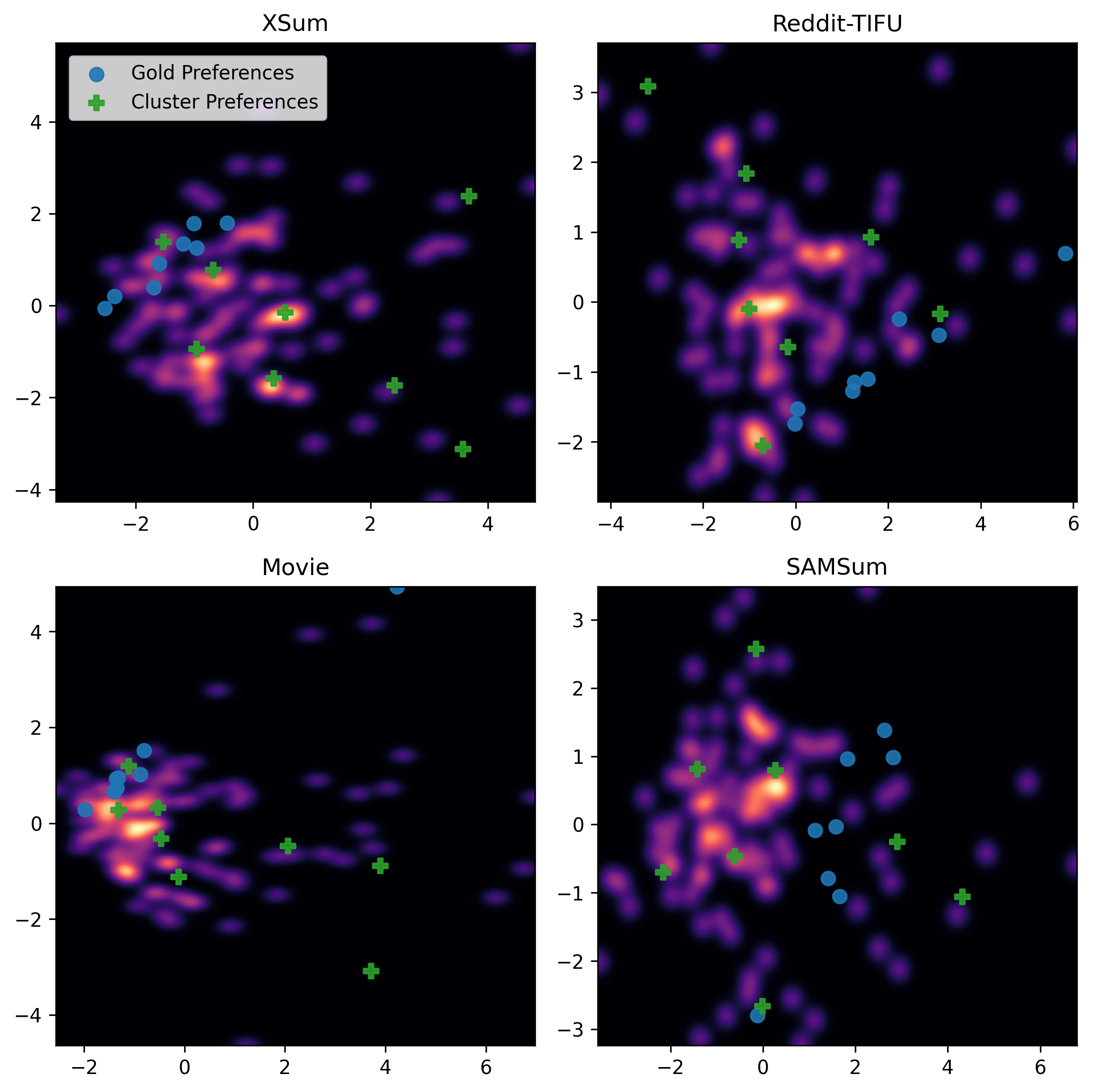}
\caption{Visualization of preference density with gold and representative (cluster) preferences.}
\label{fig:preference_density}
\end{figure}

In the following, we further visualize this phenomenon on several corpora. Specifically, we take 100 preferences from the training set of each corpus and perform principal component analysis (PCA) to reduce the dimensions from 7 to 2. Next, we conduct kernel density estimation (KDE) using Gaussian kernels, and the density maps together with representative and selected gold preferences (also through PCA) are shown in Fig.~\ref{fig:preference_density}. By using the K-Means algorithm, there are 8 representative preferences obtained from the 100 preferences. Also, there are 8 gold preferences selected at the top-75\% average L1 distance with the 100 preferences, which means that there are around 25\% of gold preferences farther from the 100 preferences. Fig.~\ref{fig:preference_density} indicates that representative preferences in different corpora are generally located in regions with higher density (brighter in color), while gold preferences are located in regions with a lower density. As such, models could generate summaries with better quality in the high-density regions since there are more demonstrations from the training examples. The preference-match decoding explores these high-density regions through the representative preferences during inference, thus improving over using the gold preferences.

From another perspective, Fig.~\ref{fg:token_loss} shows the token negative log-likelihood of using gold and representative preferences. The results show that the losses from preference-match decoding are more concentrated than those with gold preferences, which suggests that preference-match decoding can help generations become precise. In contrast, generations with gold preferences tend to distribute token losses, consistent with previous studies indicating that maximum likelihood estimation can produce high-recall models that distribute the overall loss~\cite{pang2021text}. However, typical evaluations, such as ROUGE, require models to be high-precision. Preference-match decoding creates such properties to improve the performance further.

\subsection{Generation Cases}
\label{subsec:generation_case}
Table~\ref{tb:generation_examples} demonstrates the generations on an unseen XSum document with different representative preferences. The results indicate that \TASLPabbrev~can generate diverse summaries, and the resulted preferences are quite close to the assigned preferences comparing to each other. However, the resulted ROUGE-2 scores have a wide range from 0 to 62.5. Moreover, we can find that preferences 1, 2, and 5 actually convey the same idea as the gold summary. The preference 1 and 5 use word \textit{"reinstate"} and \textit{"come back"} that are semantically identical to \textit{"relaunch"}, and they also use additional adverb clauses to provide more information without changing the meaning. While the three generated summaries are all valid, preference 5 provides the closest result to the gold summary. Considering preferences 2 and 4 with ROUGE-2 scores of 26.67 and 0, the sentences are actually fluent and complete in semantics. However, the content does not focus on the restoration of the Pirate Bay website but describes the actions taken by police, which puts emphasis on different viewpoints as compared to the gold summary. The examples demonstrate our claim for introducing preference and show that \TASLPabbrev~can achieve excellent results with a higher generality. 

\begin{table*}[t]
\scriptsize
\renewcommand\arraystretch{1.0}
\caption{Generation examples on XSum with representative preferences. Preferences are shown in the order of ROUGE-1, ROUGE-2, ROUGE-L, compression ratio, extractive coverage, extractive density, and novel word ratio.}\smallskip
\label{tb:generation_examples}
\centering
\begin{tabular}{p{0.97\textwidth}}
\toprule
\textbf{Article:} A counter at thepiratebay.se showed a countdown to 1 February, but it appears to have come back online a day early.
The website, which provides links to pirated content, was taken offline following a raid in Sweden in December.
Police officers seized servers in Stockholm after a complaint was filed by a group called the Rights Alliance, which targets internet crime.
The police operation took place in an area in Nacka, south-east of Stockholm, with the area's cold weather used as a natural cooling system for computer servers.
The site was taken down in 2006 after another raid by police but reappeared online three days later.
The Pirate Bay is one of the internet's most-visited websites, and the film, music and software industries blame it for losses running into billions of pounds.
Internet service providers (ISPs) in the UK were ordered by the High Court to block access to the site in 2012.
In October Pirate Bay co-founder Gottfrid Warg was sentenced to three-and-a-half years in prison for hacking into computers and illegally downloading files.
Another co-founder, 35-year-old Peter Sunde, was arrested in Sweden last year after two years on the run and was sentenced to eight months in prison for violating copyright laws.
Meanwhile a third co-founder, Hans Fredrik Lennart Neij (known to hackers as TiAMO), was arrested while trying to cross into Thailand from Laos in November.
A message from "Winston" on the newly-relaunched site reads: "So, first we ditched the trackers. We even got rid of the torrents. Then we left the servers to enter the clouds.
"Now, we're about to take the biggest step in our history." \\
\midrule
\textbf{Gold Summary:} The Pirate Bay website has been relaunched. \\
\textbf{Gold Preference: 0.04/0.01/0.02/0.02/0.57/1.43/0.14} \\ 
\midrule
\textbf{Generated Summary given Representative \#1:} A website called The Pirate Bay has been reinstated after a police raid in Sweden. \\
\textbf{ROUGE-1/2/L:} 54.55/30.00/45.45 \\
\textbf{Assigned Preference:}  0.12/0.03/0.05/0.05/0.77/1.43/0.21 \\
\textbf{Resulted Preference:} 0.12/0.04/0.06/0.05/0.8/1.73/0.07 \\
\midrule
\textbf{Generated Summary given Representative \#2:} A website called the Pirate Bay has come back online. \\
\textbf{ROUGE-1/2/L:} 58.82/26.67/47.06 \\
\textbf{Assigned Preference:} 0.09/0.32/0.04/0.03/0.85/3.32/0.15 \\ 
\textbf{Resulted Preference:} 0.09/0.03/0.04/0.04/0.09/1.9/0.0 \\
\midrule
\textbf{Generated Summary given Representative \#3:} A Swedish police officer has raided a Pirate Bay website and seized thousands of files, including pirated music, films, music and video games. \\
\textbf{ROUGE-1/2/L:} 26.67/14.29/20.00 \\
\textbf{Assigned Preference:} 0.09/0.01/0.05/0.08/0.44/0.51/0.46 \\
\textbf{Resulted Preference:} 0.12/0.01/0.06/0.08/0.61/0.78/0.35 \\
\midrule
\textbf{Generated Summary given Representative \#4:} Police raided servers at a Swedish website and seized thousands of copies of pirated content. \\
\textbf{ROUGE-1/2/L:} 18.18/0.0/9.09 \\
\textbf{Assigned Preference:} 0.11/0.02/0.05/0.05/0.75/1.01/0.22 \\
\textbf{Resulted Preference:} 0.10/0.01/0.04/0.05/0.73/0.87/0.27 \\
\midrule
\textbf{Generated Summary given Representative \#5:} The Pirate Bay website has been reinstated after a police raid. \\
\textbf{ROUGE-1/2/L:} 66.67/62.5/66.67 \\
\textbf{Assigned Preference:} 0.08/0.03/0.04/0.04/0.75/2.29/0.22 \\
\textbf{Resulted Preference:} 0.08/0.02/0.04/0.04/0.72/1.45/0.09 \\
\bottomrule
\end{tabular}
\end{table*}

\section{Conclusion}
In this paper, we investigate several problems of abstractive summarization under low-resource settings. To tackle these problems, we first propose a meta-transfer learning framework named \AAAIabbrev. \AAAIabbrev~leverages adapters and diverse corpora to combine the advantages of transfer and meta learning. Also, we provide empirical rules to construct meta-datasets for better generalization on target corpora. Next, we propose \TASLPabbrev~that decomposes the contents and preferences through modulating the adapter parameters with features of preferences. \TASLPabbrev~further leverages both in-domain and out-domain examples under the preference-aware architecture. Moreover, \TASLPabbrev~utilizes the preference-match decoding to automatically generate suitable summary candidates. Experimental results demonstrate that \AAAIabbrev~outperforms self-supervised baselines with 6.31\%/16.36\%/7.36\% average improvements on ROUGE-1/2/L under 10-example settings, and \TASLPabbrev~further improves \AAAIabbrev~to achieve state-of-the-art performance on six diverse corpora with 30.11\%/33.95\%/27.51\% and 26.74\%/31.14\%/24.48\% average improvements under 10- and 100-example settings. Ablation studies show that decoupling preferences from contents could help models avoid negative transfer problems, and the preference-match decoding could provide suitable summary candidates with limited supervision. In the future, we plan to explore methods for better leveraging distant source corpora, which is crucial since there is no guarantee for the availability.


%





\ifCLASSOPTIONcaptionsoff
  \newpage
\fi



%



\bibliographystyle{IEEEtran}
\bibliography{main.bib}

%

\begin{IEEEbiography}[{\includegraphics[width=1in,height=1.25in,clip,keepaspectratio]{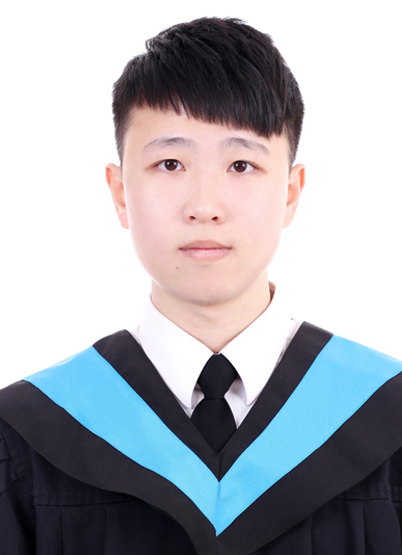}}]{Yi-Syuan Chen}
received a B.S. degree from the Department of Electrical and Computer Engineering and M.S degrees from the Institute of Electronics at National Chiao Tung University (NCTU). He is currently working toward a Ph.D degree in the Department of Electronics and Electrical Engineering, National Yang Ming Chiao Tuning University (NYCU). His research interests include natural language generation, transfer and meta learning.
\end{IEEEbiography}

\begin{IEEEbiography}[{\includegraphics[width=1in,height=1.25in,clip,keepaspectratio]{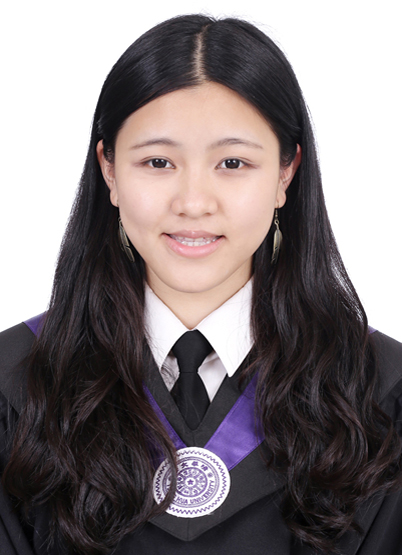}}]{Yun-Zhu Song}
received a B.S. degree from the Interdisciplinary Program of Engineering at National Tsing Hua University (NTHU). She is currently working toward a Ph.D degree in the Department of Electronics and Electrical Engineering, National Yang Ming Chiao Tuning University (NYCU). Her research interests include emotion-aware language generation, adversarial text generation and rumor detection.
\end{IEEEbiography}

\begin{IEEEbiography}[{\includegraphics[width=1in,height=1.25in,clip,keepaspectratio]{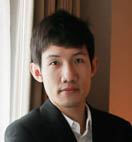}}]{Hong-Han Shuai} received a B.S. degree from the Department of Electrical Engineering, National Taiwan University (NTU), Taipei, Taiwan, R.O.C., in  2007, an M.S. degree in computer  science from NTU in 2009, and a Ph.D. degree from the Graduate Institute of Communication Engineering, NTU, in 2015. He is now an associate professor in NYCU. His research interests are in the areas of multimedia processing, machine learning, social network analysis, and data mining. His works have appeared in top-tier conferences such as MM, CVPR, AAAI, KDD, WWW, ICCV, ICDM, CIKM and VLDB, and top-tier journals such as TKDE, TMM and JIOT. He is a member of IEEE.
\end{IEEEbiography}




\end{document}